%% file: top_arxiv.tex
\documentclass[10pt,journal,compsoc]{IEEEtran}
\ifCLASSOPTIONcompsoc
  \usepackage[nocompress]{cite}
\else
  \usepackage{cite}
\fi
\ifCLASSINFOpdf
\else
\fi

\usepackage{times}
\usepackage{epsfig}
\usepackage{graphicx}
\usepackage{amsmath}
\usepackage{amssymb}
\usepackage{makecell}

\usepackage[linesnumbered,ruled,vlined]{algorithm2e}
\SetKwInput{KwInput}{Input}                % Set the Input
\SetKwInput{KwOutput}{Output}              % set the Output

\SetCommentSty{mycommfont}

\usepackage{amsfonts}
\usepackage[table]{xcolor}
\usepackage{float}
\usepackage{booktabs}
\usepackage{multirow}
\usepackage{cuted}
\usepackage{capt-of}
\usepackage{xcolor}
\usepackage{bbm}
\usepackage{pifont}%

\usepackage[pagebackref=true,breaklinks=true,letterpaper=true,colorlinks,bookmarks=false]{hyperref}

\input{defs}
\begin{document}
\title{Temporally-Consistent Surface Reconstruction using Metrically-Consistent Atlases}

\author{
	Jan~Bednarik, Noam~Aigerman, Vladimir~G.~Kim, Siddhartha~Chaudhuri, Shaifali~Parashar, Mathieu~Salzmann,~\IEEEmembership{Member,~IEEE,} and Pascal~Fua,~\IEEEmembership{Fellow,~IEEE}\IEEEcompsocitemizethanks{\IEEEcompsocthanksitem J. Bednarik, S. Parashar, M. Salzmann and P. Fua are with CVLab, School of Computer and Communication Sciences, EPFL. E-mail: see www.epfl.ch/labs/cvlab/people. \IEEEcompsocthanksitem N. Aigerman, V. G. Kim and S. Chaudhuri are with Adobe Research. S. Chaudhuri is also with IIT Bombay. Email: see research.adobe.com/people}
}

\IEEEtitleabstractindextext{%
\input{sections/abstract.tex}

\begin{IEEEkeywords}
deformable surface, temporal consistency, surface reconstruction, atlas-based representation, unsupervised shape correspondence
\end{IEEEkeywords}}

\maketitle

\IEEEdisplaynontitleabstractindextext

% \fi
\IEEEpeerreviewmaketitle

% Sections.
\input{sections/introduction.tex}
\input{sections/related_work}
\input{sections/methodology}
\input{sections/experiments}
\input{sections/conclusion}

\input{sections/acknowledgments.tex}

%% use section* for acknowledgment
%\ifCLASSOPTIONcompsoc
%  % The Computer Society usually uses the plural form
%\else
%  % regular IEEE prefers the singular form
%\fi


\ifCLASSOPTIONcaptionsoff
  \newpage
\fi



%\bibliographystyle{IEEEtran}

\bibliographystyle{ieee}
\bibliography{string,vision,graphics,learning}

\newpage ~ \newpage
\input{sections/supplementary.tex}

\end{document}

%% file: defs.tex
\newif\ifdraft
\draftfalse
\drafttrue

\definecolor{orange}{rgb}{1,0.5,0}
\definecolor{magenta}{rgb}{1,0,1}
\definecolor{cyan}{rgb}{0,0.8,0.8}
\definecolor{green}{rgb}{0,0.8,0}
\definecolor{yellow}{rgb}{0.8,0.7,0.1}

\ifdraft
 \newcommand{\JB}[1]{{\color{blue}{\bf JB: #1}}}
 
 \newcommand{\NA}[1]{{\color{orange}{\bf NA: #1}}}
 
 \newcommand{\VK}[1]{{\color{magenta}{\bf VK: #1}}}

 \newcommand{\PF}[1]{{\color{red}{\bf PF: #1}}}
 
 \newcommand{\MS}[1]{{\color{green}{\bf MS: #1}}}

\else
 \newcommand{\JB}[1]{}
 
 \newcommand{\NA}[1]{}
 
 \newcommand{\VK}[1]{}
 
 \newcommand{\SC}[1]{}
 \newcommand{\sc}[1]{ #1 }
 \newcommand{\PF}[1]{}
 
 \newcommand{\MS}[1]{}
 
\fi

\newcommand{\comment}[1]{}
\newcommand{\parag}[1]{\vspace{1mm}\textbf{#1}}

\newcommand{\real}{\mathbb{R}}

\newcommand{\floor}[1]{\left\lfloor #1 \right\rfloor}

\newcommand{\uvdom}{\Omega}

\newcommand{\lossmc}{\mathcal{L_{\text{metric}}}}
\newcommand{\lossfit}{\mathcal{L_{\text{cnstr}}}} 
\newcommand{\losscd}{\mathcal{L}_{CD}}
\newcommand{\lossrot}{\mathcal{L_{\text{rigid}}}}

\newcommand{\alphamc}{\alpha_{\text{mc}}}
\newcommand{\alpharot}{\alpha_{\text{rg}}}

\newcommand{\nricp}{nrICP}
\newcommand{\atlasnet}{AN}
\newcommand{\dsr}{DSR}
\newcommand{\cyccon}{CC}

\newcommand{\ours}{OUR}
\newcommand{\oursnolrot}{OUR w/o $\lossrot$}

\newcommand{\mdist}{m_{sL2}}
\newcommand{\mrank}{m_{r}}
\newcommand{\mpckauc}{m_{\text{AUC}}}
\newcommand{\mcd}{CD}

\newcommand{\anim}{ANIM}
\newcommand{\animr}{ANIMr}
\newcommand{\ama}{AMA}
\newcommand{\amaa}{AMAa}
\newcommand{\amaan}{AMAan}
\newcommand{\dfaust}{DFAUST}
\newcommand{\cape}{CAPE}
\newcommand{\inria}{INRIA}
\newcommand{\cmu}{CMU}

\newcommand{\cmark}{\ding{51}}%
\newcommand{\xmark}{\ding{55}}%

\def\poincare/{Poincar\'e}

\newcommand{\nrm}[1]{\left\Vert#1\right\Vert}

\newcommand{\parr}[1]{\left (#1\right )}

\def\mobius/{M{\"o}bius}

%% file: sections/abstract.tex
\begin{abstract}	

We propose a method for unsupervised reconstruction of a temporally-consistent sequence of surfaces from a sequence of time-evolving point clouds. It yields dense and semantically meaningful correspondences between frames. We represent the reconstructed surfaces as atlases computed by a neural network, which enables us to establish correspondences between frames. The key to making these correspondences semantically meaningful is to guarantee that the metric tensors computed at corresponding points are as similar as possible. We have devised an optimization strategy that makes our method robust to noise and global motions, without {\it a priori} correspondences or pre-alignment steps. As a result, our approach outperforms state-of-the-art ones on several challenging datasets. The code is available at \url{https://github.com/bednarikjan/temporally_coherent_surface_reconstruction}.

\end{abstract}

%% file: sections/introduction.tex
\begin{strip}\centering
	\vspace{-1.5cm}
	\includegraphics[width=\textwidth]{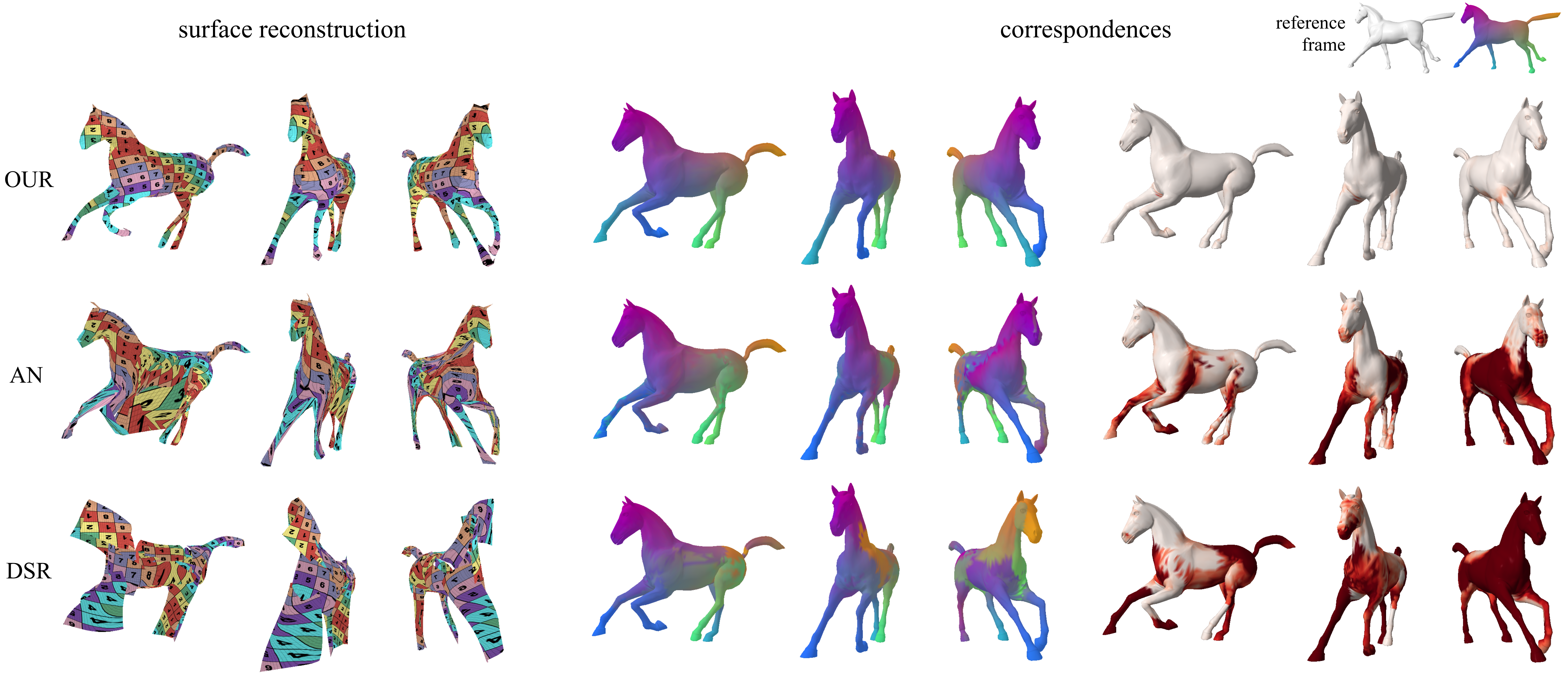}
	\captionof{figure}{{\bf Temporally consistent reconstruction and correspondences predicted by our approach (\ours{}), compared with other atlas-based methods, AtlasNet \cite{Groueix18a} (\atlasnet{}) and Differential Surface Representation \cite{Bednarik20a} (\dsr{}).} Left: The reconstructed surfaces, textured with a consistent texture to exhibit the temporally-consistent correspondences between the surfaces. Middle and right: Deviations visualized using a colormap (middle) and a heatmap (right). The competing methods exhibit artifacts and wrong correspondences, especially when the deforming object strays away from its original orientation, while OUR yields reconstructions close to the GT regardless of the object's transformation.}
	\label{fig:teaser}
	\vspace{0.5cm}
\end{strip}

\IEEEraisesectionheading{\section{Introduction}\label{sec:introduction}}

\IEEEPARstart{A}{pplications} such as UV-mapping, shape analysis, and partial scan completion all rely on having a surface representation that remains consistent over time. Specifically, the different surfaces reconstructed at different times should be relatable to each other so that each point on one surface maps to a point with the same semantic meaning on another. The most common way to achieve this is to explicitly establish correspondences between potentially inconsistent representations, such as 3D meshes~\cite{Varanasi10,Arcila13,Roufosse19,Halimi19,Donati20,Rakotosaona20} or 3D point clouds~\cite{Insafutdinov18,Groueix19}.

This, however, assumes that the input data contains points that can be matched in a semantically-meaningful manner and does not enforce true consistency of the representations. In this paper, we tackle this problem more directly by learning to reconstruct temporally-consistent surfaces from a sequence of 3D point clouds representing a shape deforming over time. To this end, we rely on the AtlasNet patch-based representation~\cite{Groueix18a} to model the surface underlying the 3D points. However, whereas in the original AtlasNet, any patch can describe any part of the surface, we enforce consistency of the patch locations across the whole sequence. Hence, we create a time-consistent atlas in which each 2D point on each 2D patch models the same semantic 3D point over time by leveraging differential geometry. Specifically, when the deformations are isometric, the metric tensor computed at any surface point remains constant even when the shape changes. We translate this into a metric consistency loss function, which, when minimized, implicitly establishes the desired point correspondences. Even when the deformation is not strictly isometric, the deformation which real-world surfaces undergo from one instant to the next is sufficiently close to being isometric so that imposing isometry as a soft constraint over short time intervals is a valid approximation.

Because we use differential properties instead of ground-truth correspondences, our approach is unsupervised and can operate on any shape category {\it without} a known shape template. Yet, as shown in Fig.~\ref{fig:teaser}, it provides reliable correspondences even when the shapes are complex and the deformations severe, unlike state-of-the-art methods that tend to break down. 

We first introduced this approach in a conference paper~\cite{Bednarik21}. This early version yielded state-of-the-art results in terms of the predicted correspondences but required a data pre-alignment step to handle global object rotations. In this paper, we remove this limitation by introducing an additional self-supervised correspondence loss term and a novel progressive data sampling strategy that jointly enable our algorithm to handle rotating objects without any pre-processing, which the SotA methods cannot do. We demonstrate this on several synthetic and real-world datasets. Furthermore, we test the robustness of our method on datasets with noise of various severity and show that it outperforms all the SotA methods with large margin.

%% file: sections/related_work.tex
\section{Related Work}

3D temporal consistency involves both surface reconstruction and correspondence estimation. Both are essential geometry processing tasks and we briefly review the existing literature below. 

\subsection{Correspondence Estimation} 
Estimating correspondences between instances  of a non-rigidly deforming object is a well-studied problem. We start with reviewing the traditional, mostly optimization based methods, then we move on to the more recent ones relying on deep learning.

The traditional approaches explicitly focus on either \textit{isometric} or \textit{non-isometric} deformations. The isometry well approximates deformation of common real-world soft-tissue objects such as human bodies or animals. Majority of isometry-based methods rely on availability of triangulated meshes which, in contrast to point clouds, represent topologically unambiguous surfaces. Typically, local shape descriptors are computed~\cite{Ovsjanikov12,Aubry11,Sun09c,Li13e} and used to establish correspondences among pairs of shapes without need for GT annotations. Other methods enforce local distance preservation~\cite{Bronstein06,Baden18} or assume piece-wise rigidity of the target shapes but only deliver region-wise correspondences~\cite{Arcila13,Varanasi10}. Yet another popular approach is to parameterize the input 3D mesh models into a common 2D base-domain in which correspondences are established~\cite{Aigerman14,Weber14}, but some form of GT annotations is needed. Our approach uses 2D domain to define correspondences as well, however, it works directly on the input point clouds and it keeps the preset 2D correspondences fixed while optimizing the 3D surface to perform shape reconstruction.

Far fewer methods which estimate correspondences directly on the point clouds exist. \cite{Cuzzolin08} clusters local linear embeddings of the input data to produce only part segmentation without dense correspondences. \cite{Memoli05} aligns isometrically deformed shapes by analyzing geodesic distances computed on point clouds which, however, break down when the shape topology become ambiguous (e.g. a human bringing their arms close to their body).

The methods which support non-isometric deformations either explicitly model deviation from isometry~\cite{Lipman09}, rely on conformal deformation~\cite{Baden18,Lipman10,Koehl14,Le16,Wang05e} or make no assumption about the deformation type~\cite{Kraevoy04,Rodola14,Loper15,Zuffi15} whatsoever, which allows them to align objects under extreme deformations (e.g. a cat and a dog). When meshes are available, a popular approach is to conformally map a pair of shapes to a common 2D domain in the form of a disc or a 2-sphere and find the alignment there using M{\"o}bius transformation. This can be done with the help of a-priory known corresponding landmarks~\cite{Kraevoy04,Wang05e} or in an unsupervised way~\cite{Lipman09,Lipman10,Koehl14,Le16,Baden18}. Not only do these methods require the availability of meshes but in the absence of GT landmarks they only support disc-like or genus 0 surface topology. A different approach was introduced in~\cite{Rodola14} where correspondence estimation is cast as a classification problem with the use of random forest, but supervision is needed. Yet another group of approaches assume the knowledge of a shape template~\cite{Loper15,Zuffi15} or a close enough shape initialization~\cite{Cosmo19} and are thus restricted to a given class of shapes. In contrast, our method does not need any template and rather than conformal deformations it assumes isometry which is well constrained and provides reliable estimate on correspondences of common real-world deforming objects (e.g. a dancing human).

The most recent methods make use of deep learning paradigm which allows for arbitrary capacity of the learned mappings and for very fast inference times. The setting of isometry on meshes gained the most traction. For example, the functional maps of~\cite{Ovsjanikov12} have been reformulated in a differentiable way to be incorporated into the deep learning pipeline~\cite{Halimi19,Donati20,Roufosse19}.

Once again, point clouds have been dealt with much less. \cite{Groueix18b} assumes isometry and uses a human shape template which restricts it to a single object class. The most recent methods~\cite{Niemeyer19b,Tang21} represent arbitrary shapes via an implicit function and do not assume any specific deformation type, however, they require GT correspondences. Finally, \cite{Insafutdinov18,Groueix19} are the only unsupervised template-free methods working directly on point clouds. \cite{Insafutdinov18} only yields a set of semantically close points without a mechanism to find a unique correspondence rendering the method unsuitable for the task of dense shape correspondences estimation. On the other hand, \cite{Groueix19} learns correspondences by enforcing cyclic consistency across multiple shape-triplets and can retrieve dense correspondences. However, it tends to break down under more complex articulated motion and/or under severe global object rotation.

In summary, contrary to all the aforementioned methods, ours yields temporally-consistent surface reconstructions from point clouds and generates meaningful point-wise correspondences, without either template or supervision.  

\subsection{Surface Reconstruction}
Triangulated meshes have long been used for 3D surface reconstruction~\cite{Fua96f,Kanazawa18b,Pan19} but obtaining a mesh is not always practical and furthermore, these methods tend to require fixed topology which severely restricts their applicability to practical scenarios. By contrast, implicit fields such as signed distance functions and occupancy grids~\cite{Park19c,Mescheder19} can handle complex and changing topologies but their implicit formalism does not naturally lend itself to establishing correspondences. 

Many other mathematical tools have been developed for surface reconstruction from point clouds. They include  solving the Poisson PDE~\cite{Kazhdan13} or using Moving Least Squares~\cite{Lipman07} to fit points to the surface. We refer the interested reader to~\cite{Berger14} for a survey. Recently, deep learning techniques have been successfully used to reconstruct 3D shapes as point clouds~\cite{Qi17a,Qi17b,Fan17a,Insafutdinov18}.
While compact and flexible, point clouds lack the ability to represent a complete surface without topological ambiguities.
Therefore, parametric functions that represent surfaces as a 2D manifolds embedded in 3D space were introduced in a deep learning context. It started with the seminal FoldingNet~\cite{Yang18a}, AtlasNet~\cite{Groueix18a}, and their followups~\cite{Deprelle19,Williams19a,Deng20b}. In earlier work~\cite{Bednarik20a}, we showed that the metric tensor could be computed analytically by differentiating the learned parametric mappings and used either to control the amount distortion the mappings undergo. In this work, we show that imposing metric consistency not only results in temporal consistency but also in visually better reconstructions.
  
\subsection{Metric Preservation and Shape Interpolation} 

Our approach is closely related to shape interpolation because it can be formulated as a metric preservation problem in which a low-distortion map between shapes is sought. Shape interpolation has long been studied in computer graphics~\cite{Aaron99,Alexa00b,Winkler10}. In recent years, several data-driven methods have been proposed for this task~\cite{Gao17b,Cosmo20}, but they assume to be given point  correspondences and do not infer them. The method of~\cite{Rakotosaona20} is designed to smoothly interpolate between two point clouds, without correspondences being given {\it a priori}.  This makes it close in spirit to ours but it focuses solely on interpolating the point clouds without generating meaningful correspondences nor producing a continuous surface.

%% file: sections/methodology.tex
\section{Methodology}

\subsection{Problem Statement and Overview}

Let us assume we are given as input  a temporal sequence of 3D point clouds $P_1,...,P_K$. The desired output is a corresponding sequence of reconstructed surfaces $\mathcal{S}_1,...,\mathcal{S}_K$, one per point cloud, along with a canonical bijective mapping $\Psi_{i,j}$ between the surfaces $S_i,S_j$ that defines temporally-consistent point-to-point correspondences. To this end, we use an atlas-based representation with multiple patches as in~\cite{Groueix18a}.  Each atlas ${\Phi}_j$ represents a surface $\mathcal{S}_j$. This naturally defines a canonical bijective map $\Psi_{i,j}$ between any two surfaces $\mathcal{S}_i$ and $\mathcal{S}_j$ because their atlases share the same 2D domain, as shown in Fig.~\ref{fig:metric_consistency_schema}. We wish to optimize the atlases so that the corresponding surfaces satisfy the following two properties.
\begin{enumerate}

    \item \textbf{Goodness of fit.} Each surface $\mathcal{S}_k$ represents the corresponding point cloud $P_k$ as faithfully as possible.
    
    \item \textbf{Temporal consistency.} Each $\Psi_{i,j}$ maps semantic parts of the surface consistently across frames so that semantic features correspond to each other.  
    
\end{enumerate}
Our core intuition is that we can  achieve this goal in an unsupervised manner by making the $\Psi_{i,j}$ as isometric as possible, thus encouraging the transition from one frame in the sequence to the next to preserve local shape features, thereby making the reconstructions consistent. In practice, even when the deformation is not strictly isometric, it remains nearly so over neighboring time frames and our approach provides the right kind of regularization.

%%%%%%%%%%%%%%%%%%%%%%%%%%%%%%%%%%%%%%%%%%%%%%%%%%%%%%%%%%%%%%%%%%%%%%%%%%%%%%%%
\subsection{Atlas-Based Point Cloud Representation}

\input{fig/atlas}

In its simplest form, an \emph{atlas} can be defined as a map $\phi:\uvdom\to \mathbb{R}^3$ from a 2D domain $\uvdom$ to a 3D surface $\mathcal{S}$, such that the image of $\phi$ is $\mathcal{S}$. In practice, we take $\uvdom$ to be $[0, 1]^2$ in all our experiments. Such atlases let us define a canonical point-to-point correspondence between any two 3D surfaces, $\mathcal{S}_1$ and $\mathcal{S}_2$, described by atlases, $\phi_1$ and $\phi_2$ respectively. Given a point $p \in \uvdom$, $\phi_1\left(p\right)\in \mathcal{S}_1$ corresponds to $\phi_2\left(p\right)\in \mathcal{S}_2$, which we denote by $\Psi$ in  Fig.~\ref{fig:metric_consistency_schema}. 

In practice, we use AtlasNet~\cite{Groueix18b} to implement our atlases. A Multi Layer Perceptron takes as input a point $p\in\uvdom$ and a latent code $z \in \real^{C}$, where $C$ is the dimension of the latent space, and outputs a 3D point. Since AtlasNet relies on $P\geq1$ patches, it defines for each surface $\mathcal{S}_k$ 
%several mappings $\phi^1_i,\phi^2_k,...,\phi^P_k$ 
several mappings $\phi^1_i, 1 \leq i \leq P$, each mapping the same point $p \in \uvdom$ to $\phi^{i}_{k}(p) \in \mathcal{S}_{k}$ individually. As this holds for any $P$, for simplicity, we take $\phi_k$ to be the aggregate mapping, that is an ensemble of $P$ mappings.

To reconstruct each surface independently, we train AtlasNet to minimize the reconstruction loss 
\begin{small}
\begin{align}
\lossfit &= \frac{1}{K}\sum_{k=1}^K  \mathcal{L}_{\text{CD}}^k \; , \label{eq:lossFit} \\
\mathcal{L}_{\text{CD}}^k &= \int_{p \in \Omega}{\min_{q \in P_k}{||\phi_k(p) - q||^{2}}} + \sum_{q \in P_k}{\min_{p \in \Omega}{||\phi_k(p) - q||^{2}}} \; \label{eq:cd} ,
\end{align}
\end{small}
where $\mathcal{L}_{\text{CD}}^k$ is the symmetric Chamfer distance to point cloud $P_k$. This yields reconstructions that may be accurate but does not guarantee that they are consistent across point clouds. For that, we need an additional consistency term, which we describe below.

%%%%%%%%%%%%%%%%%%%%%%%%%%%%%%%%%%%%%%%%%%%%%%%%%%%%%%%%%%%%%%%%%%%%%%%%%%%%%%%%
\subsection{Enforcing Consistency across Time}
\label{sec:timeConsistency}

Key to enforcing consistency is the observation that when a surface deforms isometrically, the {\it Riemannian metric tensor} remains unchanged at any given 3D point on the surface. In other words, with $g_{\phi_k}(p)$ the  metric tensor at point $\phi_k(p)$ of $\mathcal{S}_k$, for any two surfaces  $\mathcal{S}_{k_1}$ and $\mathcal{S}_{k_2}$ we must have
\begin{equation}
\forall p \in \Omega \quad g_{\phi_{k_1}}(p) = g_{\phi_{k_2}}(p) 
\label{eq:constTensor}
\end{equation}
if $\phi_{k_1}(p)$ and $\phi_{k_2}(p)$ map to the same 3D point, which they should if the two mappings are consistent. 

The metric tensor can be computed as
\begin{align}
g_{\phi}(p) &= J_\phi\left(p\right)^{\top}{J_\phi\left(p\right)} \; , \label{eq:metricTensor} \\
J_{\phi}(p) & =\left[\frac{\partial \phi(p)}{\partial u}, \frac{\partial \phi(p)}{\partial v}\right] \; \nonumber,
\end{align}
where $J_\phi\in\mathbb{R}^{3\times2}$ is the matrix of partial derivatives of the map $\phi$ at $p$. Given $g$ and two arbitrary vectors  $q,r\in\mathbb{R}^2$, we can define their local inner product as $q^T g_\phi \left(p\right) r$, which in turns allows us to estimate local lengths and angles at $p$.

When a surface deforms isometrically, the  metric tensor remains unchanged. Hence, given two surfaces $\mathcal{S}_1$ and $\mathcal{S}_2$ and their respective  mappings, $\phi_1$ and $\phi_2$, for all $p \in \uvdom$, $g_{\phi_1}(p)$ and $g_{\phi_2}(p)$ should be equal. In practice, the deformation is not necessarily truly isometric and Eq.~\ref{eq:constTensor} is not truly satisfied. We therefore define the {\it metric consistency loss}
 \begin{align}
     \lossmc &= \sum_{\left(i,j\right)\in \mathcal{I}} E_\text{cons}\parr{\phi_i,\phi_j},     \label{eq:consLoss} \\
    E_\text{cons}\parr{\phi_1,\phi_2} &= \int_{p\in\uvdom} \nrm{g_{\phi_1}\parr{p}-g_{\phi_2}\parr{p}}^2_F \; ,
\end{align}
where $\mathcal{I}$ is a set of surface pairs and $\| \cdot \|_F$ is the Frobenius norm. Minimizing $E_\text{cons}\parr{\phi_1,\phi_2}$ then enforces preservation of the metric tensor across corresponding points. 

Assuming that the shape gradually deforms over time and, thus, that surfaces in nearby frames deform almost isometrically, we define a time window $\delta$ and only consider pairs of frames that fall within this window. In other words, we take the set $\mathcal{I}$ in Eq.~\ref{eq:consLoss} to be $\{ (i,j) | \left|i - j \right| \leq \delta \}$.

%%%%%%%%%%%%%%%%%%%%%%%%%%%%%%%%%%%%%%%%%%%%%%%%%%%%%%%%%%%%%%%%%%%%%%%%%%%%%%%%
\subsection{Equivariance to Global Rotations and Translations}
\label{sec:rotations}

In our earlier work~\cite{Bednarik21}, we trained the AtlasNet by minimizing a weighted sum of $\lossfit$ and $\lossmc$ over the complete sequence. However, we found out experimentally that this does not suffice when the object or parts of it undergo a substantial global rotation and translation throughout the sequence. The patches have a tendency to creep across the surface as it rotates and/or translates with respect to their locations in neighboring frames. To prevent this, we introduce the data augmentation scheme and additional loss term discussed below. 

\input{fig/augment}

To force the model to adjust appropriately the orientation of each mapping $\phi_{k}$, we want to expose the model not only to the original point clouds but also to rotated and translated versions. To this end, for each $P_k$, we select $O$ rotations by sampling random rotation axes $u$ as shown in Fig.~\ref{fig:augm}(a). For each $o \in O$, we produce a point cloud  $P_{k}^o$ rotated by $R_{o}$ and translated by $T_{o}$. For each $p$ in $\Omega$, we then associate to $\phi_k(p)$ the closest point in $P_{k}$, which we denote by $P_{k}(\phi_{k}(p))$. We then introduce the new loss function
\begin{align}
\lossrot = &\frac{1}{O}\sum_{o=1}^{O}\int_{p \in \Omega}|| R_{o}{P_{k}}(\phi_{k}(p)) + T_{o} - \phi_{k}^o(p) ||^{2} \; ,
\label{eq:lossRot}
\end{align}
where $\phi_{k}^{o}$ is the mapping produced by our model for $P_{k}^{o}$. Minimizing this loss means that the mapping $\phi_{k}^o$ should map $p$ to a point that is close to the rotated version of the point closest to $\phi_{k}(p)$. In other words, it should be equivariant to rotations and translations. 

%%%%%%%%%%%%%%%%%%%%%%%%%%%%%%%%%%%%%%%%%%%%%%%%%%%%%%%%%%%%%%%%%%%%%%%%%%%%%%%%
\subsection{Complete Loss Function}

We then take our full loss function to be
\begin{equation}
    \mathcal{L} = \lossfit + \alphamc \mathcal{L}_\text{metric} + \alpharot \lossrot \;, 
    \label{eq:fullLoss}
\end{equation}
where $\alphamc \in \real$ and $\alpharot \in \real$ are hyper-parameters of our approach balancing the individual loss terms. Note that $\lossfit$ and $\lossmc$ are computed for the original point clouds $P_{k}$ only while $\lossrot$ is computed only for the transformed ones $P_{k}^{o}$.

%%%%%%%%%%%%%%%%%%%%%%%%%%%%%%%%%%%%%%%%%%%%%%%%%%%%%%%%%%%%%%%%%%%%%%%%%%%%%%%%
\subsection{Implementation Details}
\label{sec:implementation}

\subsubsection{Optimization}

We use a slightly modified version of AtlasNet~\cite{Groueix18a} as in~\cite{Bednarik20a}: ReLU is replaced by Softplus in the decoder; and the batch normalization layers are removed to make the convergence more stable. We systematically use $P = 10$ patches, the Adam optimizer with a learning rate $l = 0.001$, and a batch size of $4$ for $200000$ iterations. We introduce a learning rate scheduler which divides the current $l$ by a factor of $10$ at $80\%$ and $90\%$ of the training iterations. As in~\cite{Groueix18b,Bednarik20a}, $2500$ points are sampled from the UV domain $\uvdom$. We set both weights $\alphamc$ and $ \alpharot$ in Eq.~\ref{eq:fullLoss} to $0.1$, which, as we will see in the ablation study of the results section, yields good results across all datasets.  We set $\delta$ for each dataset individually using one validation sequence and test on all others using that value. We provide an ablation study of $\delta$ in the appendix. 

\subsubsection{Progressive Sequence Sampling} 
\label{sec:progressive_data_sampling}

As discussed in Section \ref{sec:rotations}, we introduce the loss term $\lossrot$ to handle global rotations and translations. However, when the target object presents strong geometric symmetries---front and back of a male human body or left and right side of an animal----the model can still get confused. Even though, during the optimization, we sample pairs of point clouds that are close in time, as explained in Section \ref{sec:timeConsistency}, the pairs are still sampled from the whole sequence and consecutive iterations may feature samples whose global orientation is very different but nevertheless look the same because of the symmetries. 

To counteract this, we rely on progressive sequence sampling. Instead of sampling point cloud pairs from the whole sequence, we start with a small time window in the middle of it and progressively expand it to cover the whole sequence. This allows the model to correctly track the surface of a potentially rotating object.

Specifically, given a sequence of $K$ point clouds,  we start optimizing the model using only the $5$ middle ones $[{\floor{K/2} -2},{\floor{K/2} + 2}]$ and optimize for $I_{\text{init}}$ iterations. This bootstraps the model and allows the mappings $\phi_{k}$ to spatially expand so as to cover the whole reconstructed surface. Then, between iterations $I_{\text{init}}$ and $I_{\text{end}}$ we gradually and symmetrically expand the time window by a single frame at a time until the full sequence is used for optimization. We set $I_{\text{init}} = 30000, I_{\text{end}} = 150000$ in all our experiments.

%% file: fig/atlas.tex
\begin{figure}[tb]
\begin{center}
\includegraphics[width=0.50\textwidth]{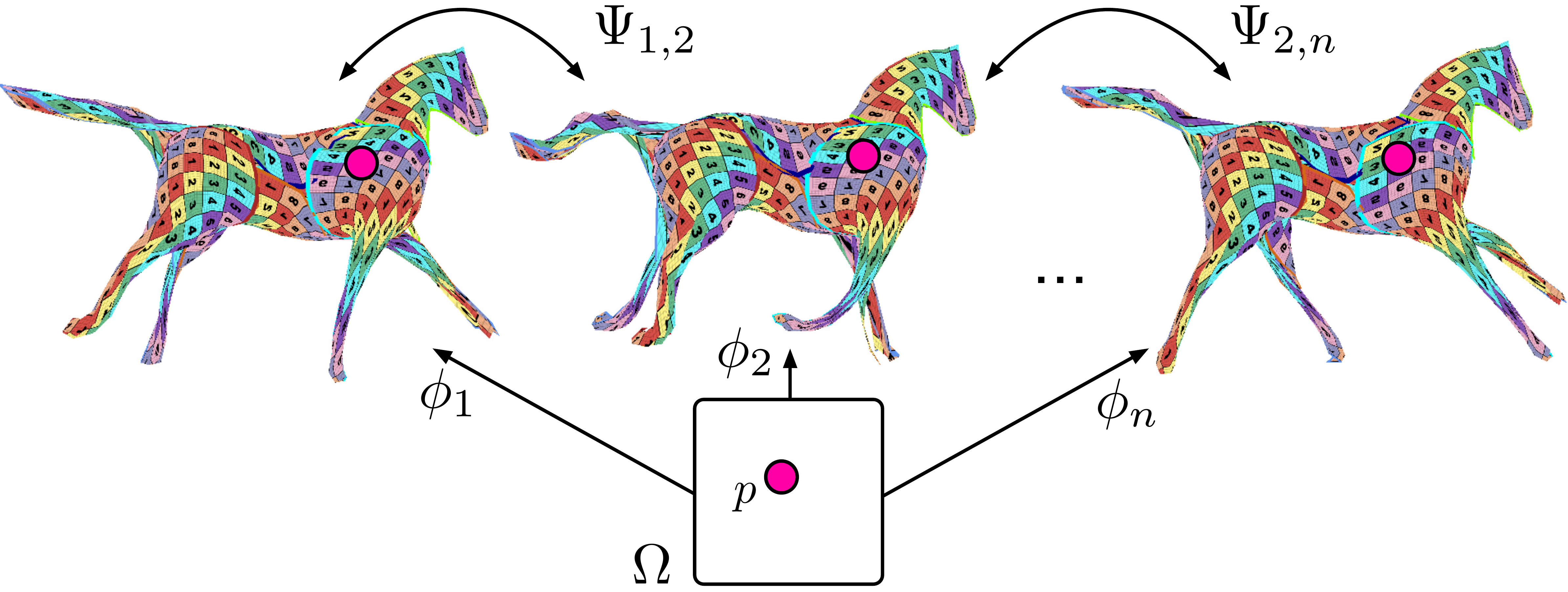}
\end{center}
  \caption{{\bf Atlas and correspondences.} Correspondences defined between three surfaces by the mapping of one point $p\in\Omega$ through three different atlases. }
\label{fig:metric_consistency_schema}
\end{figure}

%% file: fig/augment.tex
\begin{figure}[th]
	\begin{center}
		\includegraphics[width=0.99\linewidth]{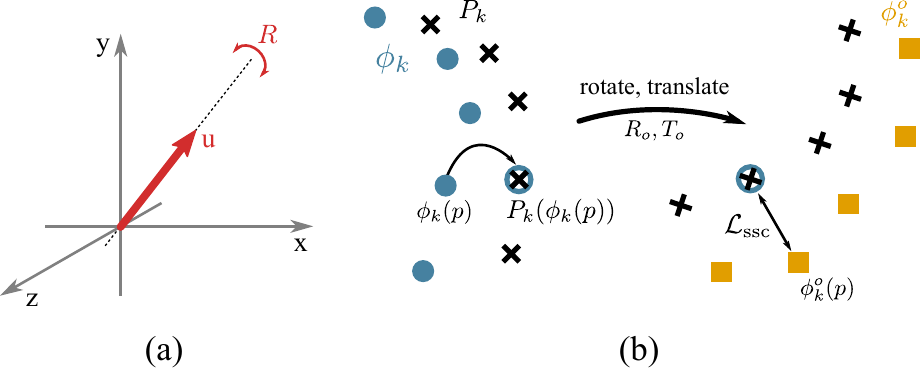}
	\end{center}
	\caption{{\bf Data Augmentation.}{\bf (a)}  Randomly sampled rotation axis. {\bf (b)} The original and rotated/translated point cloud $P_{k}$ along with the corresponding reconstructions.}
	\label{fig:augm}
\end{figure}

%% file: sections/experiments.tex
\section{Experiments} 
\label{sec:experiments}

\begin{figure*}[th]
\begin{center}
\includegraphics[width=0.99\linewidth]{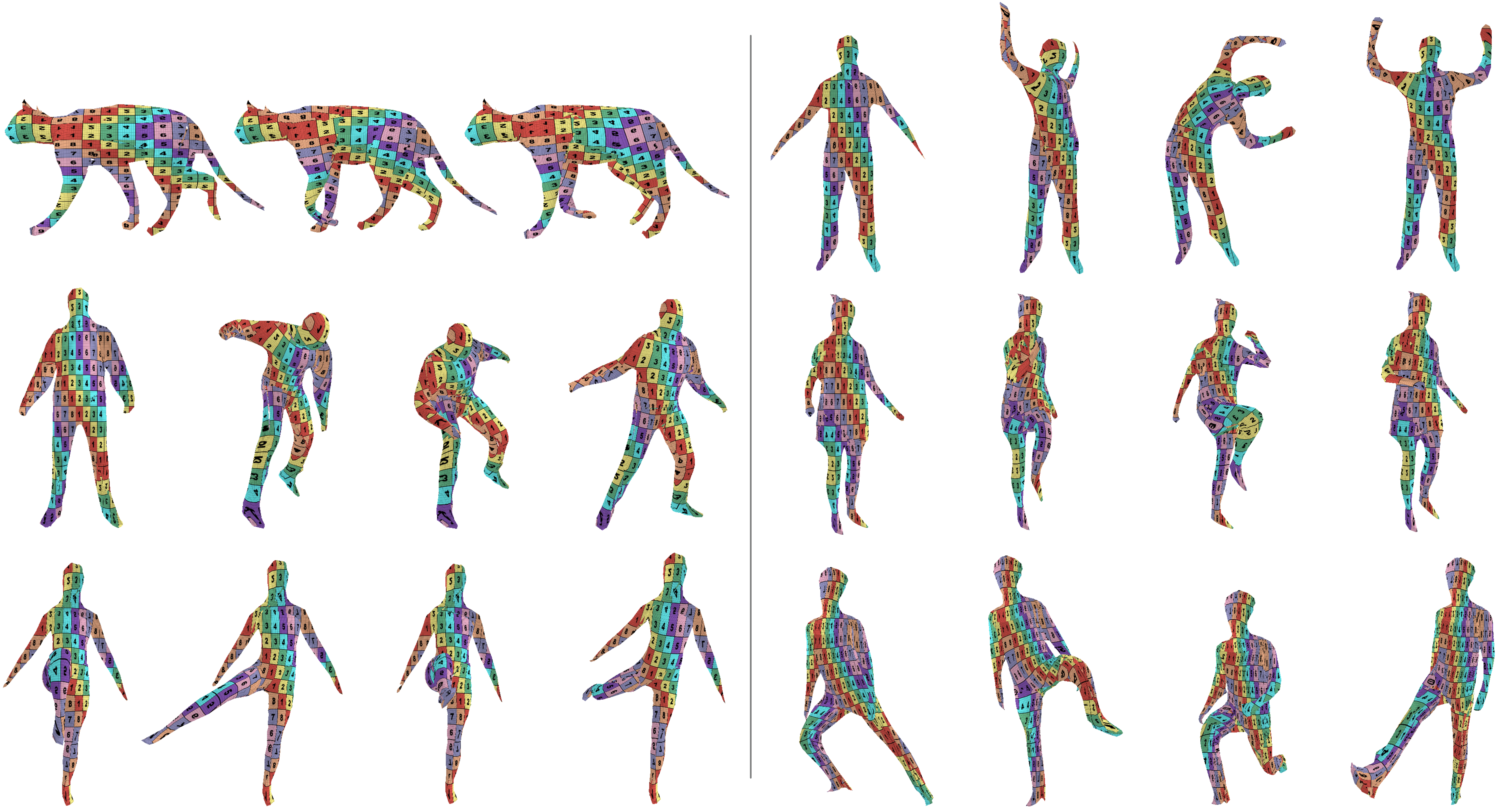}
\end{center}
  \caption{\textbf{Our temporally-coherent surface reconstructions, for 6 sequences.} Left: \texttt{cat} (\anim{}), \texttt{jumping} (\ama{}) and \texttt{knees} (\dfaust{}), right: \texttt{shortlong\_tilt\_twist\_left} (\cape{}), \texttt{wide\_knee} (\inria{}) and \texttt{low\_lunges} (\cmu{}). Note how the reconstructed surfaces have consistent correspondences, as well as accurate geometry.}
\label{fig:reconstructions_our}
\end{figure*}

To demonstrate that our approach can handle many types of objects without supervision, known correspondences, or any kind of template, we test it on several point-cloud sequences featuring human and animal motions. Some of them are clean while others are much noisier or feature global object motion. We also provide videos of the reconstructed sequences as supplementary material. 

%%%%%%%%%%%%%%%%%%%%%%%%%%%%%%%%%%%%%%%%%%%%%%%%%%%%%%%%%%%%%%%%%%%%%%%%%%%%%%%%
\subsection{Datasets}\label{sec:datasets}
We evaluate our method on nine datasets. To generate point clouds from the meshes or triangle soups they contain, we sample randomly and uniformly $2500$ points from each frame.

\begin{figure}[th]
	\begin{center}
		\includegraphics[width=0.99\linewidth]{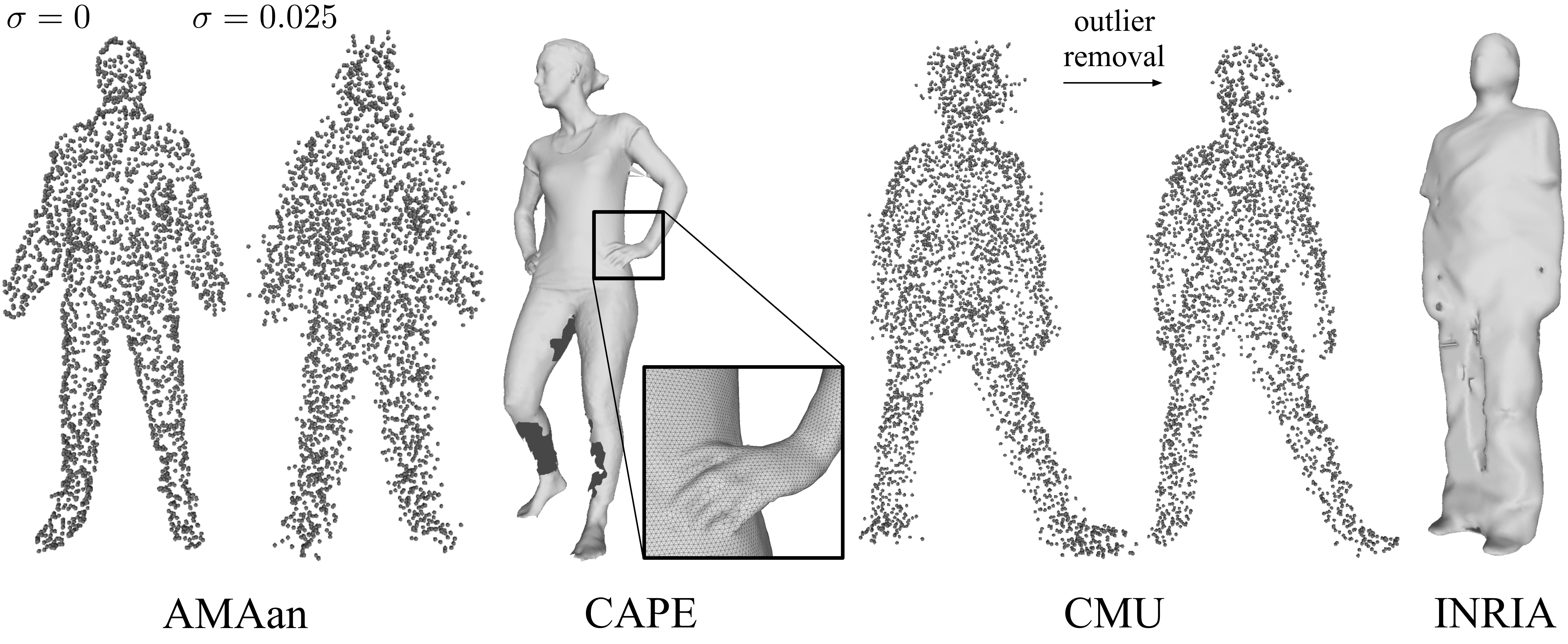}
	\end{center}
	\caption{\textbf{Noisy datasets.} \amaan{} adds a spatial Gaussian noise on top of AMA. The triangle soups of \cape{} exhibit wrong topology, severe holes and spurious geometry. The real-world point clouds of \cmu{} contain strong noise which we attenuate by automatic statistical outlier removal. The meshes of \inria{} suffer from wrong topology, especially when the human limbs are close to the body or to each other.}
	\label{fig:noisy_datasets}
\end{figure}

\parag{Animals in motion \cite{Sumner04,Aujay07} (\anim{})} comprises 4 synthetic mesh sequences depicting strides of four-legged animals. We uniformly scaled each sequence so that the first point cloud fits in a unit cube. To test the influence of global rotations, we created a more challenging variant, \textbf{\animr{}}, in which each animal is gradually rotated around the vertical axis up to the angle of $180^{\circ}$ so as to mimic running in a circle.

\parag{Dynamic FAUST \cite{Bogo17} (\dfaust{})} is a real-world dataset which contains $14$ mesh sequences of $10$ unclothed human subjects performing various actions without any global rotation. We sample the point clouds from the clean registered meshes.

\parag{Articulated mesh animation \cite{Vlasic08} (\ama{})} is a real-world dataset containing $10$ mesh sequences depicting $3$ different humans performing various actions. The subjects wear loose-fitting clothes making the surfaces more complex and time-varying. Furthermore, it features large rotations such as when people walk in a circle. We sample the point clouds from the registered meshes. To test our approach under scenarios of varying difficulty, we introduced two variants. In \textbf{\amaa}, we eliminated the global rotations by finding in each frame the angle around the main rotation axis that minimizes the chamfer distance with respect to the the previous frame. In \textbf{\amaan{}}, we added a substantial amount of Gaussian noise---$\sigma = 0.025$---to the  \amaa{} point clouds, as shown in Fig. \ref{fig:noisy_datasets}.

\parag{Clothed auto person encoding \cite{Ma20} (\cape{}) } is a real-world dataset of humans wearing loose-fitting clothing and performing various actions, without any global rotation. We use the raw scans containing $4$ subjects performing $8$ different motions. We preprocess the data by thresholding away the ground and selecting the biggest connected component from the remaining triangle soup. While comparable to \dfaust{}, this dataset features substantially more noise coming from the wrong topology and/or spurious geometry, as can be seen in Fig.~\ref{fig:noisy_datasets}. We use the associated clean registered meshes for evaluation purposes only.

\textbf{Human in Motion with Wide Clothing \cite{Yang16c} (\inria{})} is a real-world dataset featuring $4$ human subjects performing $3$ actions while wearing different types of clothing. The ground-truth data consists of $14$ sparse keypoints attached to major human body joints. We discarded the sequences of subjects wearing layered clothing because there is no ground-truth data for the garments, which  left $4$ sequences per subject. As \cape{}, the dataset is noisy due to errors in mesh topology and geometry, as shown in Fig. \ref{fig:noisy_datasets}. Some sequences feature substantial subject rotation at the end of the motion.

\parag{CMU Panoptic Studio \cite{Joo15} (\cmu{})} is a real-world dataset featuring point clouds of people performing actions both alone and in groups. The data was captured using several Microsoft Kinect depth sensors with poor time synchronization, making the resulting point clouds highly noisy. As no ground-truth registration is provided, we use this dataset for qualitative comparison only. We selected a sequence where only one person appears (\texttt{171026\_pose3}) and removed the chunks where the subject is fully static, which yielded the $4$ sequences, an example is depicted in Fig. \ref{fig:noisy_datasets}. Because the data is extremely noisy with many spurious points and objects, we performed a standard automated preprocessing in the form of point cloud clustering and statistical outlier removal.

%%%%%%%%%%%%%%%%%%%%%%%%%%%%%%%%%%%%%%%%%%%%%%%%%%%%%%%%%%%%%%%%%%%%%%%%%%%%%%%%
\subsection{Metrics}
\label{sec:metrics}

Our method can be used to infer point-to-point correspondences on the input point clouds. Given two point clouds $P_k$ and $P_l$, we can map points by Euclidean projections from the point clouds to the reconstructed surfaces. More formally, we can define the mapping $f_{k \rightarrow l} = \pi_{P_k} \circ \phi_k \circ \phi_l^{-1}  \circ  \pi_{\phi_l}$, where $\pi_\mathcal{X}$ denotes the projection of a 3D point to its nearest neighbor on $\mathcal{X}$, be it a point cloud $P$ or the image of a mapping $\phi$. In addition to be a useful application in of itself, we can also use this to evaluate the accuracy of our method given ground-truth correspondences. To this end, we randomly draw $M$ pairs of point clouds $(P_k, P_l)$ with known ground-truth correspondences $(p_{kl}^i, q_{kl}^i)$, where $p_k \in P_k$, $q_k \in P_l$, and $i$ ranges from 1 to the total number of ground-truth points.

\parag{Squared correspondence distance ($\mdist$).} It is computed as $\mdist = \frac{1}{N} \sum_{i=1}^N \| f_{k \rightarrow l}(p_{kl}^i) - q_{kl}^i\|^2$.

\parag{Normalized correspondence rank ($\mrank$).}  It it the rank of a predicted point with respect to all the other points on the target object computed as  $\mrank = \frac{1}{N^2}\sum_{i=1}^{N}\sum_{j=1}^{N}\mathbbm{1}_{\|q_{kl}^i-q_{kl}^j\|^2 < \|f(p_{kl}^i)-q_{kl}^i\|^2}$.

\parag{Area under the percentage of correct keypoints (PCK) curve (\textnormal{$\mpckauc$}).} As in the literature on keypoint classification and correspondences~\cite{Huang17f,You20}, we compute a PCK curve in a range $[d_{\text{min}}, d_{\text{max}}]$ and report the area under that curve (AUC). We set $d_{\text{min}} = 0$ and $d_{\text{max}} = 0.02$, which corresponds to the maximum real world distance of $14$ cm, in all our experiments. 

\parag{}We also report the \textbf{Chamfer Distance (CD)}, which we take to be the loss term $\losscd$ of Eq.~\ref{eq:cd}, to characterize the quality of the reconstruction. When evaluating all these metrics, we ignore any patch with an area smaller than $1/1000$ of the average patch area. We provide additional details in the appendix.

\begin{figure}[tbh]
	\begin{center}
		\includegraphics[width=0.99\linewidth]{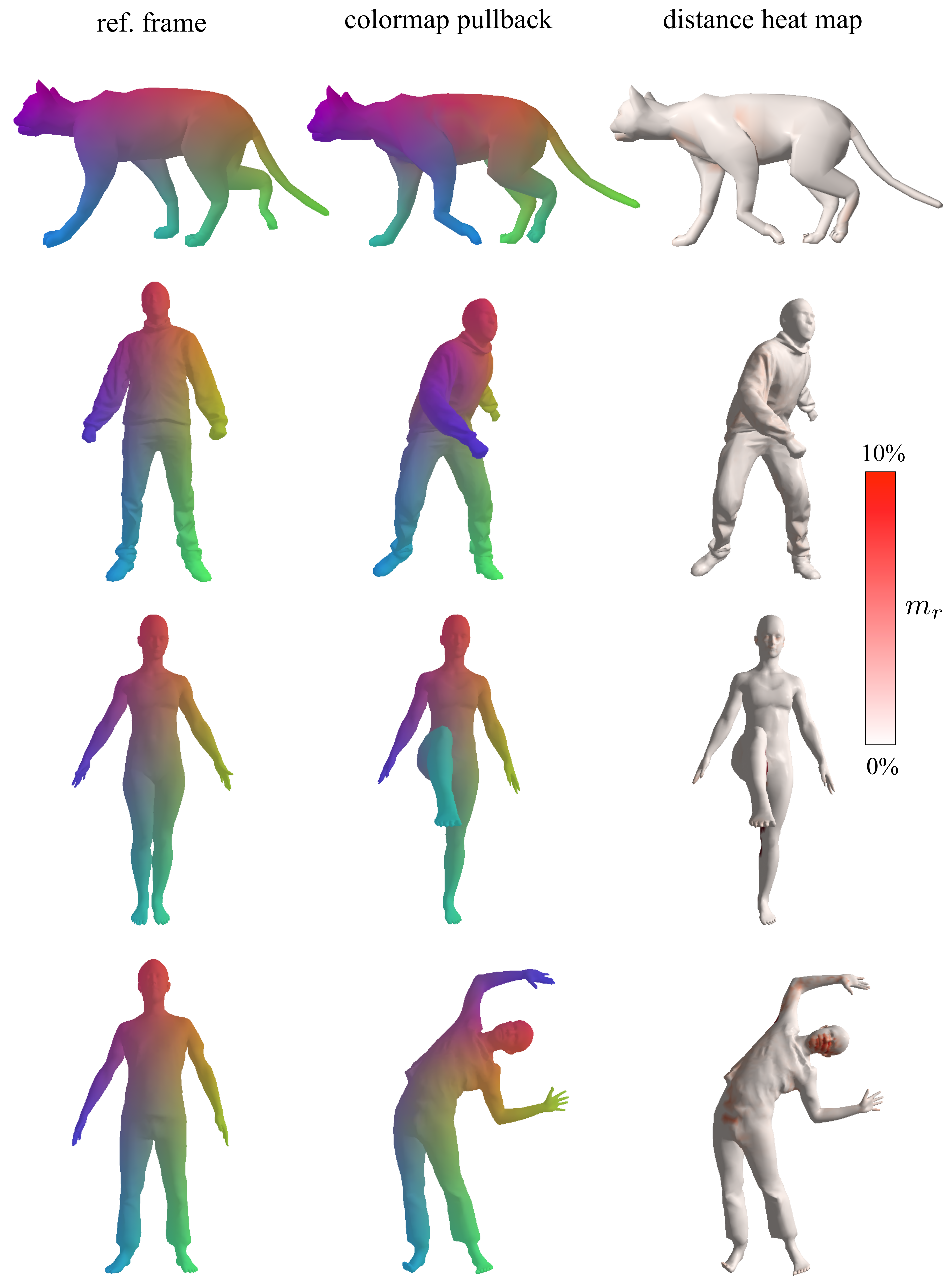}
	\end{center}
	\caption{\textbf{Our correspondences retrieved on \texttt{cat} from ANIM, \texttt{jumping} from AMA, \texttt{knees} from DFAUST and \texttt{shortlong\_tilt\_twist\_left} from \cape{}.} We visualize the correspondences via matching colors and show the error colorcoded as a heat map on the right. }
	\label{fig:correspondence_our}
\end{figure}

%%%%%%%%%%%%%%%%%%%%%%%%%%%%%%%%%%%%%%%%%%%%%%%%%%%%%%%%%%%%%%%%%%%%%%%%%%%%%%%%
\subsection{SotA Methods}
\label{sec:sota_methods}

We compare our approach (\ours{}) to both traditional and deep learning based methods. 

\parag{Non-rigid ICP.} It is a popular and classic technique for shape registration. We use the recent implementation of~\cite{Huang17f}, which we denote as \nricp{}. We experimented with several ways to use it to match shape pairs and chose the optimal one. Please refer to the appendix for details.

\parag{Atlas-based methods.} As our method builds on an atlas-based representation, we compare it to the original AtlasNet \cite{Groueix18a} (\atlasnet{}). We also compare it to our more recent method~\cite{Bednarik20a} (\dsr{}) that aims to reduce patch distortion but does not account for temporal distortion.

\parag{Cycle consistent point cloud deformation.} This is the recent method of~\cite{Groueix19} (\cyccon{}). It deforms one point cloud into another while maintaining semantic correspondence.

All the deep learning based methods, \atlasnet{}, \dsr{}, \cyccon{} and \ours{}, are trained directly on the given sequence and then evaluated on it to retrieve the correspondences. As the base architecture for \atlasnet{}, \dsr{}, and \ours{} are almost identical, we train all three methods as described in Section~\ref{sec:implementation} for a fair comparison. As the training of \cyccon{} relies on sampling triplets, we experimented extensively to find an optimal sampling technique, as described in the appendix. Finally, since \cyccon{} and \nricp{} do not reconstruct surfaces, we do not report CD for them.

\begin{figure*}[th]
	\begin{center}
		\includegraphics[width=0.99\linewidth]{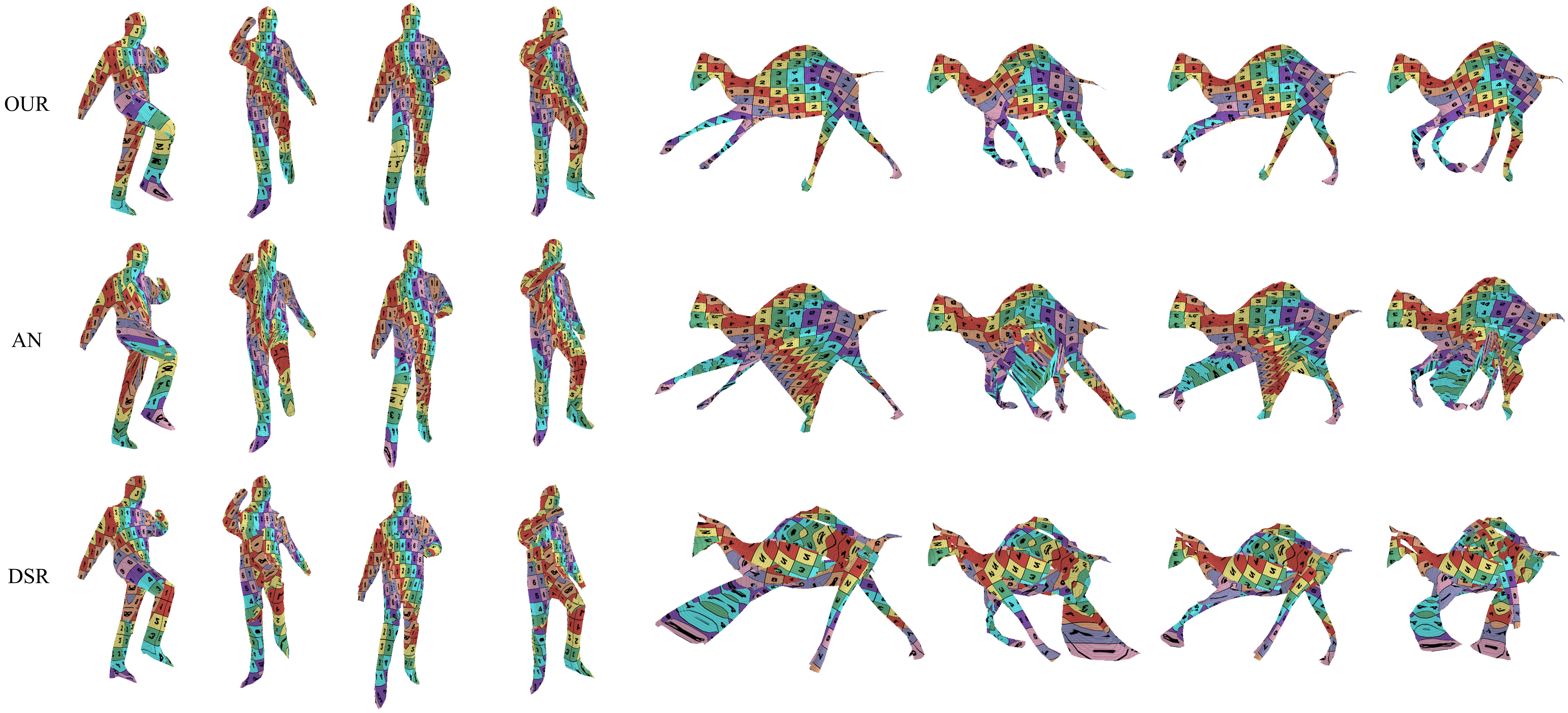}
	\end{center}
	\caption{\textbf{Comparison of the reconstructions of the data without noise produced by the atlas-based methods \atlasnet{}, \dsr{} and \ours{}.} Left: \texttt{march\_1} from \amaa{}, right: \texttt{camel\_gallop} from \anim{}. \atlasnet{} and \dsr{} struggle to reconstruct the camel and swap the legs of the human, while \ours{} does not.}
	\label{fig:comp_reconstruction}
\end{figure*}

%%%%%%%%%%%%%%%%%%%%%%%%%%%%%%%%%%%%%%%%%%%%%%%%%%%%%%%%%%%%%%%%%%%%%%%%%%%%%%%%
\subsection{Results}
\label{sec:results}
	
We trained and evaluated all methods on every sequence individually. In the case of \dfaust{}, \cape{} and \inria{} we simultaneously train on all subjects performing the sequence, but still draw point cloud pairs of the same subject.

For each dataset with available ground-truth data, we used one sequence---\texttt{cat} for \anim{}, \texttt{jumping jacks} for \dfaust{}, \texttt{crane} for \ama{}, \texttt{shortlong hips} for \cape{}, \texttt{wide knee} for \inria{}---to set the $\delta$ parameter from Section.~\ref{sec:timeConsistency} in the range 1 to 6. All three correspondence-based metrics described above are strongly correlated and we chose values that were optimal with respect to $\mdist$. We then report the metrics for the remaining sequences.

%%%%%%%%%%%%%%%%%%%%%%%%%%%%%%%%%%%%%%%%%%%%%%%%%%%%%%%%%%%%%%%%%%%%%%%%%%%%%%%%
\subsubsection{Qualitative Results} 
\label{sec:our_reconstructions_and_correspondences}

Fig.~\ref{fig:reconstructions_our} depicts our temporally-consistent reconstructions for one sequence of each dataset. In this figure, we highlight the temporal consistency of our reconstructions by using the same texture in the UV space for all frames of the sequence. Hence, corresponding regions are textured with the same checkerboard cells, which demonstrates the accuracy of the correspondences. Our algorithm successfully reconstructs high-curvature regions, such as the cat's ears and paws that are tracked as they move. It delivers both accurate geometry and correspondences. Even though the human models exhibits much more complex articulated deformations, limbs are tracked properly, and consistent, meaningful correspondences are maintained throughout the sequences.

In Fig.~\ref{fig:correspondence_our}, we use a matching colormap to visualize the correspondences produced by \ours{}. We also visualize the correspondence errors with the color denoting the magnitude of the correspondence errors. They are mostly small.

\begin{figure*}[th]
	\begin{center}
		\includegraphics[width=0.99\linewidth]{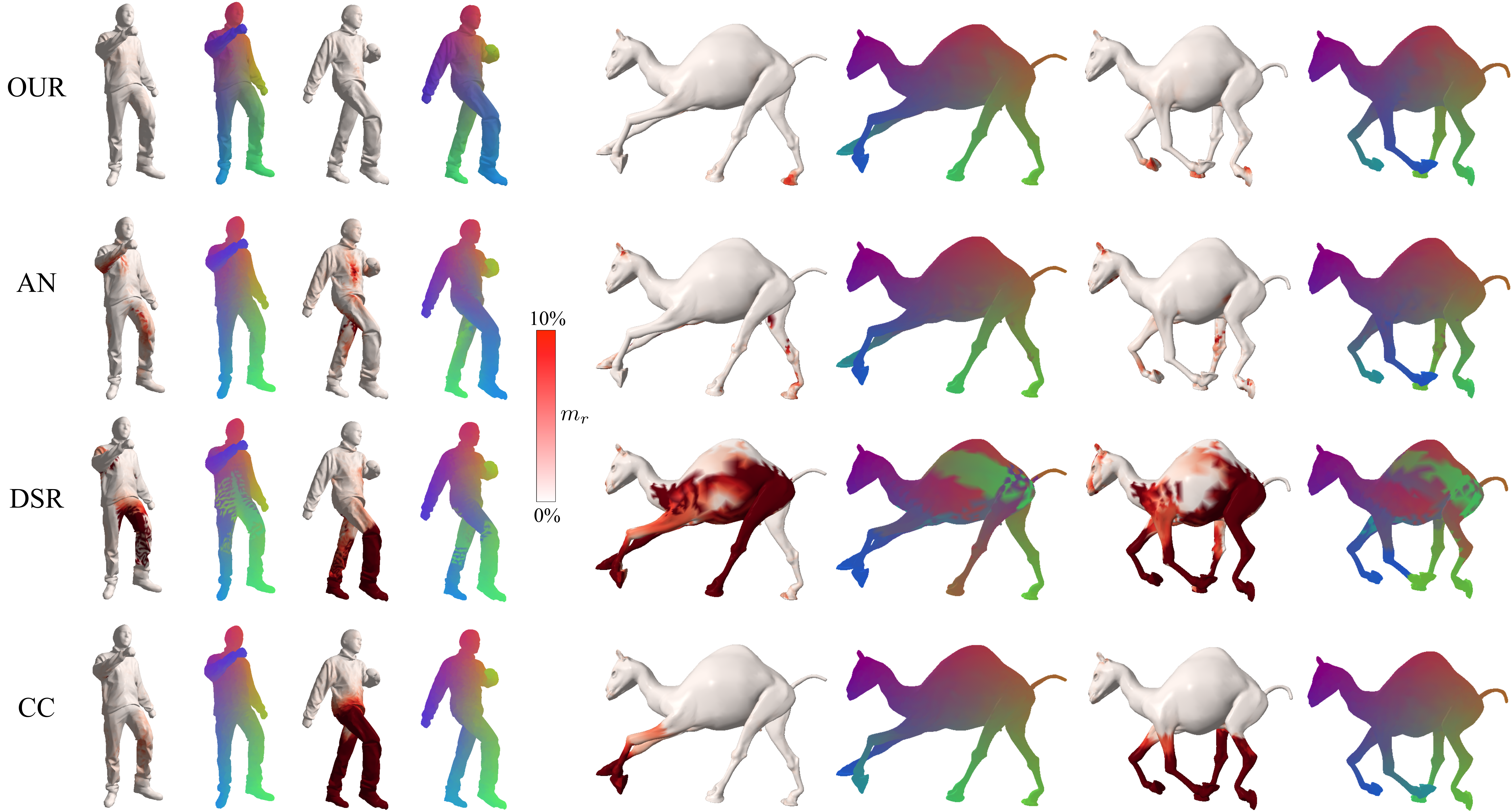}
	\end{center}
	\caption{\textbf{Comparison of inferred point-cloud correspondences} within the sequence \texttt{march\_1} from \amaa{} and \texttt{camel\_gallop} from \anim{}. Notice the correspondence errors and artifacts, such as swapped legs of the camel as predicted by \dsr{} and \cyccon{}, and minor problems, such as small but sever local mismatches, such as the knee and ears areas of the camel as predicted by \atlasnet{}.}
	\label{fig:comp_correspondence}
\end{figure*}

%%%%%%%%%%%%%%%%%%%%%%%%%%%%%%%%%%%%%%%%%%%%%%%%%%%%%%%%%%%%%%%%%%%%%%%%%%%%%%%%
\parag{Sequences without Global Rotations.} \label{sec:no_global_rotations_qualitative}
In Fig.~\ref{fig:comp_reconstruction}, we compare the reconstruction quality against that of the atlas-based baselines \atlasnet{} and \dsr{}. Our method is markedly more temporally-consistent. For example, in the leftmost frame of the human sequence, the bending at the knee causes \atlasnet{} to introduce a significant amount of unnecessary distortion while \dsr{} swaps the two legs. The camel sequence reveals something even more interesting: Temporal consistency also acts as a regularizer and makes the reconstruction itself more accurate. This is particularly noticeable on the camel's legs. 
In Fig.~\ref{fig:comp_correspondence}, we use the same visualization strategy as before to compare the correspondence accuracy of \ours{} against that of the baselines. \cyccon{} and \dsr{} tend to swap the legs of the human and the camel along the sequence while \atlasnet{} produces patches that are spatially jittering over time, which manifests as local correspondence imprecisions. See also the \href{https://youtu.be/P4imXONmtto}{supplementary video}\footnote{https://youtu.be/P4imXONmtto}.

\begin{figure*}[th]
	\begin{center}
		\includegraphics[width=0.99\linewidth]{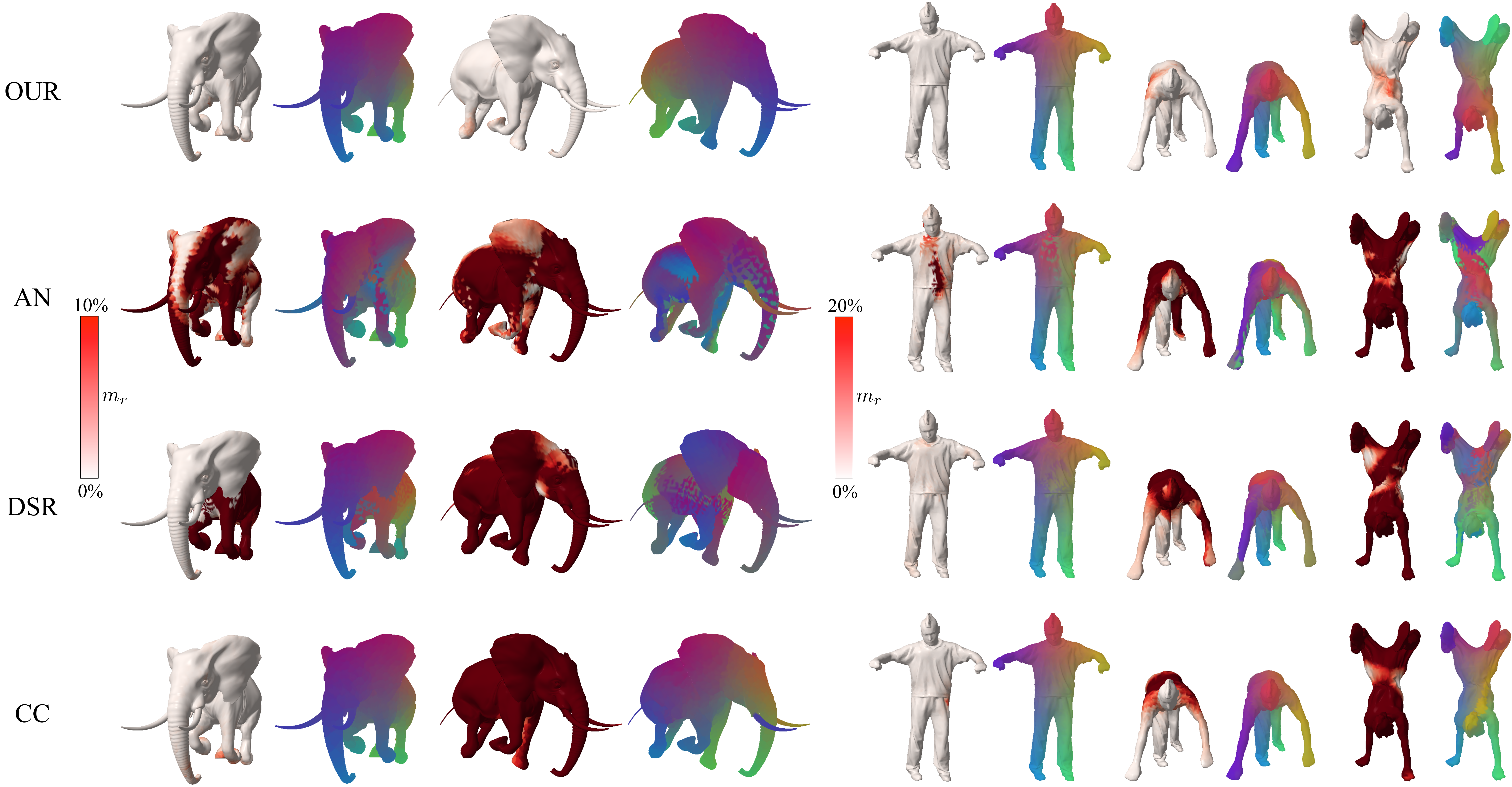}
	\end{center}
	\caption{\textbf{Comparison of inferred point-cloud correspondences on the data with global rotation} --- sequences \texttt{elephant\_gallop} from \animr{} and \texttt{handstand} from \ama{}. Once the object rotates sufficiently far away from its original orientation, none of the baselines can track the surface properly, while \ours{} infers correct correspondences regardless of the orientation.}
	\label{fig:comp_correspondence_global_tf}
\end{figure*}

%%%%%%%%%%%%%%%%%%%%%%%%%%%%%%%%%%%%%%%%%%%%%%%%%%%%%%%%%%%%%%%%%%%%%%%%%%%%%%%%
\parag{Sequences with Global Rotations.} \label{sec:global_rotations_qualitative}
In the sequences of Fig.~\ref{fig:comp_reconstruction} and \ref{fig:comp_correspondence}, the objects do not rotate much. By contrast, those showcased by Fig.~\ref{fig:comp_correspondence_global_tf} do. When using \atlasnet{} and \dsr{}, the correspondence accuracy rapidly drops as the subject rotates away from its original orientation because the individual mappings $\phi_{k}$ tend to maintain their original global position and orientation, as explained in Section~\ref{sec:rotations}. Similarly, \cyccon{} fails to recover the correct orientation of the object due to its symmetries and predicts incorrect correspondences once the object has undergone severe global rotation. Our approach suffers from none of these shortcomings.

\begin{figure*}[th]
	\begin{center}
		\includegraphics[width=0.99\linewidth]{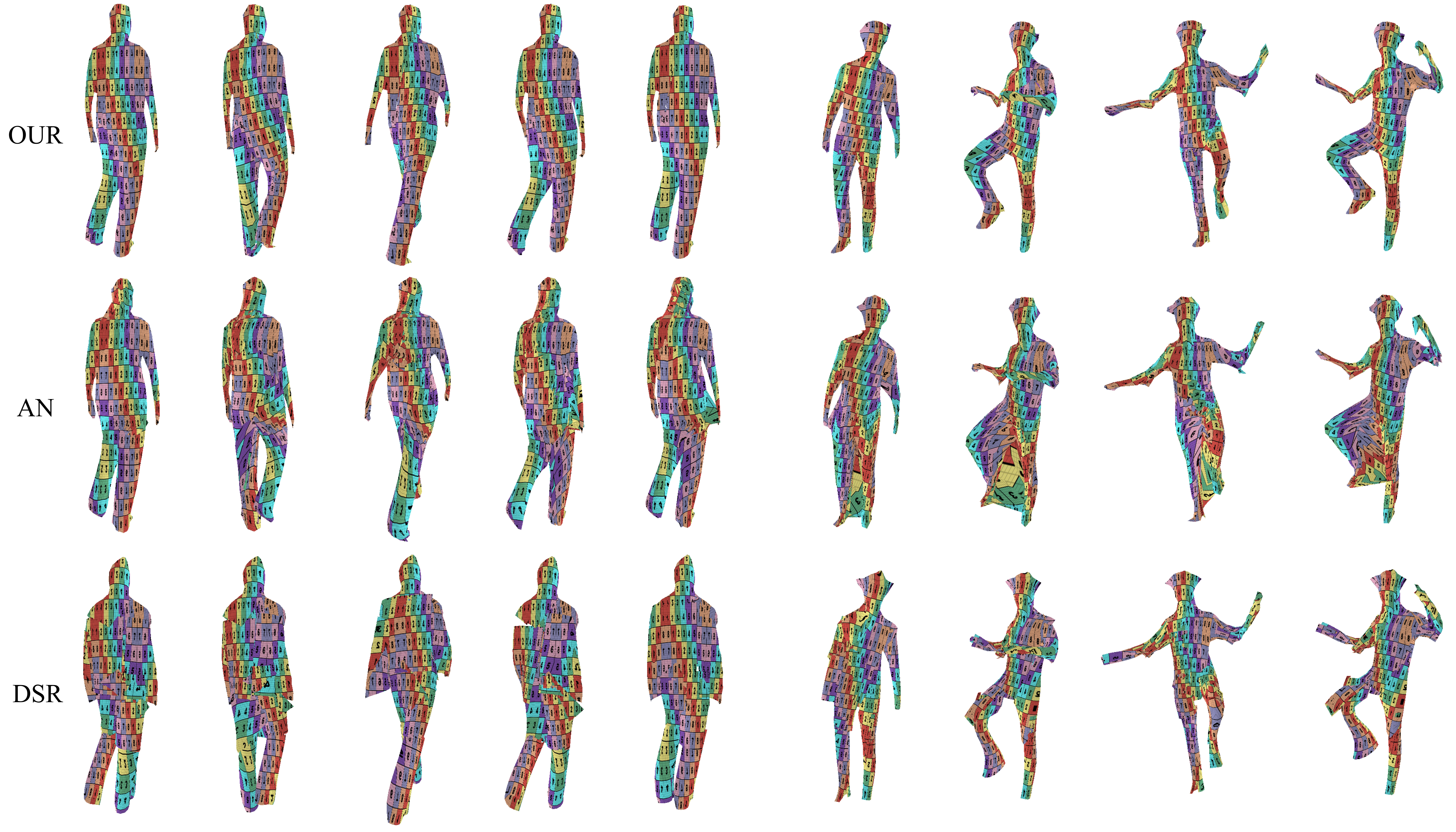}
	\end{center}
	\caption{\textbf{Comparison of the reconstructions of the noisy data as produced by the atlas-based methods \atlasnet{}, \dsr{} and \ours{}.} Left: \texttt{wide\_walk} from \inria{}, right: \texttt{arms\_fingers} from \cmu{}. \atlasnet{} and \dsr{} wrongly merge the human limbs to the body or to each other, and they often swap the human legs.}
	\label{fig:comp_reconstruction_noisy}
\end{figure*}

%%%%%%%%%%%%%%%%%%%%%%%%%%%%%%%%%%%%%%%%%%%%%%%%%%%%%%%%%%%%%%%%%%%%%%%%%%%%%%%
\parag{Sequences of Noisy Data.} \label{sec:noisy_data_qualitative}
Fig.~\ref{fig:comp_reconstruction_noisy} shows the reconstruction quality of \ours{} and the baselines on the noisy datasets \inria{} and \cmu{}. Note, that we only show the reconstructions as neither of those datasets provide GT registered meshes.

The main challenge in the \inria{} dataset stems from the fact the human subjects keep their arms close to their body during the motion (see Fig.~\ref{fig:noisy_datasets}). Both \atlasnet{} and \dsr{} often fail to discern the arms, merge them to the body and thus produce wrong correspondences. Furthermore, notice that both \atlasnet{} and \dsr{} keep swapping the human legs. The strong noise in \cmu{} causes the reconstruction quality to degrade for all the methods. However, note how \atlasnet{} struggles with identifying the individual human legs and \dsr{} produces mappings which swap between modeling the legs and the torso, while \ours{} still delivers reasonable reconstructions and correct correspondences, as further evidenced by the supplementary video.

\begin{figure*}[tbh]
	\begin{center}
		\includegraphics[width=\linewidth]{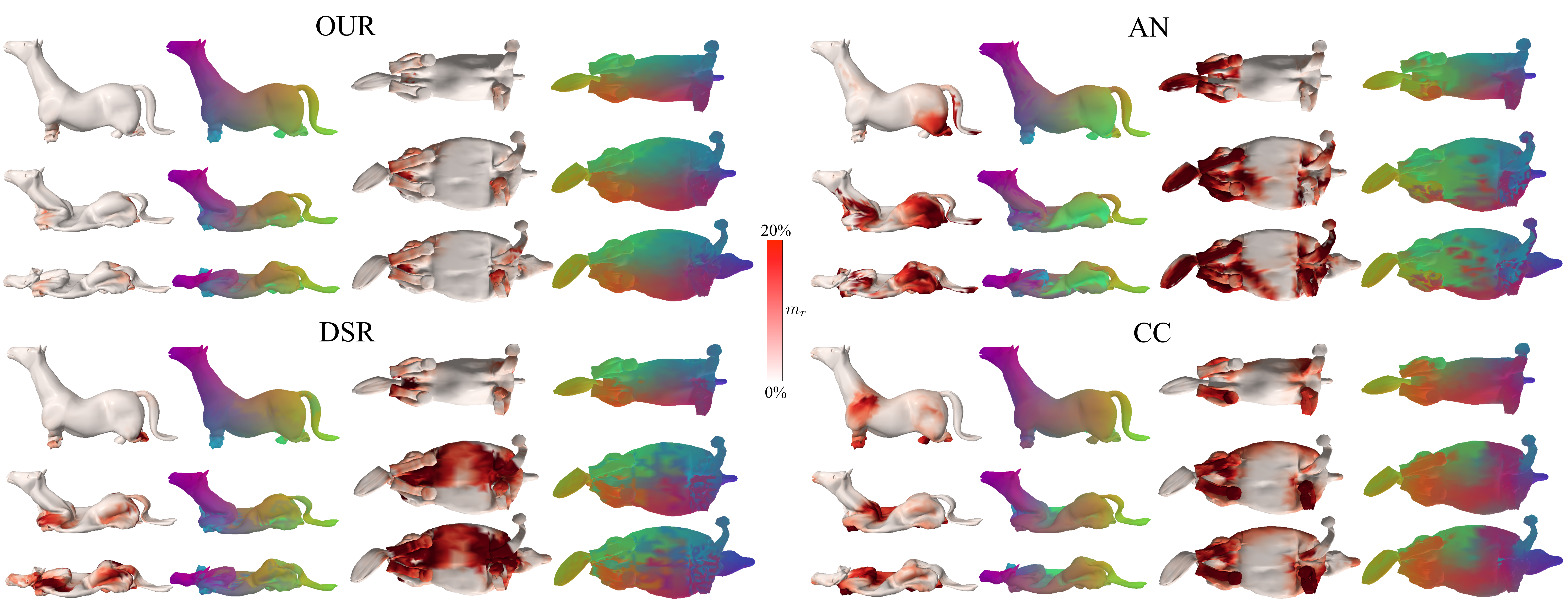}
	\end{center}
	\caption{\textbf{Stress test on the sequence \texttt{horse\_collapse} from ANIM, of a horse deflating.} Despite the many self-foldovers, our method still finds accurate correspondences. The other methods fail. For each method we show  side and bottom views of 3 frames from the sequence, with correspondence visualized via matching colors, and a heatmap showing correspondence error. }
	\label{fig:stresstest}
\end{figure*}

~\\
Finally, in Fig.~\ref{fig:stresstest}, we stress test our method on a sequence from \anim{} of a rubber horse deflating, which features extreme deformations. Even though the surface repeatedly folds over itself, our method reconstructs the legs as separate from the surface, while  the baselines clamp different regions together. This is particularly visible in the bottom rows, when the horse is almost completely deflated.

%%%%%%%%%%%%%%%%%%%%%%%%%%%%%%%%%%%%%%%%%%%%%%%%%%%%%%%%%%%%%%%%%%%%%%%%%%%%%%%%
\subsubsection{Quantitative Results}
\label{sec:quantitative_results}

\begin{table}[htbp]
  \centering
  \caption{\textbf{Comparison of \ours{} to SotA methods on correspondence accuracy and reconstruction quality using the datasets with globally aligned sequences.} Our method is the most accurate in terms of all the correspondence metrics.}
  \resizebox{0.49\textwidth}{!}{
  	\begingroup
  	\setlength{\tabcolsep}{1pt}
  	\renewcommand{\arraystretch}{1.1}
    \begin{tabular}{clcccc}
    \toprule
    \textbf{dataset} & \textbf{\phantom{sp}model\phantom{sp}} & \boldmath{}\textbf{\phantom{spac}$\mdist$ $\downarrow$\phantom{spac}}\unboldmath{} & \boldmath{}\textbf{\phantom{spa}$\mrank$ $\downarrow$\phantom{spa}}\unboldmath{} & \boldmath{}\textbf{\phantom{sp}$\mpckauc$ $\uparrow$\phantom{sp}}\unboldmath{} & \boldmath{}\textbf{\phantom{spac}CD $\downarrow$\phantom{spac}}\unboldmath{} \\
    \midrule
    \multirow{5}[2]{*}{\anim} & nrICP & 70.32$\pm$84.86 & 5.46$\pm$9.52 & 74.23$\pm$13.68 & - \\
          & AN    & 18.40$\pm$24.82 & 0.78$\pm$2.85 & 96.28$\pm$1.56 & \boldmath{}\textbf{0.09$\pm$0.00}\unboldmath{} \\
          & DSR   & 46.43$\pm$67.42 & 3.44$\pm$6.71 & 83.96$\pm$9.58 & 0.19$\pm$0.01 \\
          & CC    & 33.84$\pm$54.13 & 2.21$\pm$4.75 & 87.96$\pm$7.76 & - \\
          & \textbf{OUR} & \boldmath{}\textbf{12.60$\pm$11.38}\unboldmath{} & \boldmath{}\textbf{0.36$\pm$0.72}\unboldmath{} & \boldmath{}\textbf{98.00$\pm$0.71}\unboldmath{} & \boldmath{}\textbf{0.09$\pm$0.00}\unboldmath{} \\
    \midrule
    \multirow{5}[2]{*}{\amaa} & nrICP & 150.94$\pm$134.31 & 6.63$\pm$10.26 & 45.40$\pm$22.27 & - \\
          & AN    & 86.80$\pm$91.28 & 2.90$\pm$6.18 & 70.07$\pm$15.31 & 0.30$\pm$0.01 \\
          & DSR   & 123.56$\pm$109.92 & 5.00$\pm$7.39 & 59.69$\pm$15.94 & 62.08$\pm$52.50 \\
          & CC    & 74.58$\pm$97.98 & 2.47$\pm$6.37 & 77.07$\pm$15.00 & - \\
          & \textbf{OUR} & \boldmath{}\textbf{35.60$\pm$29.42}\unboldmath{} & \boldmath{}\textbf{0.36$\pm$0.94}\unboldmath{} & \boldmath{}\textbf{89.73$\pm$5.26}\unboldmath{} & \boldmath{}\textbf{0.27$\pm$0.01}\unboldmath{} \\
    \midrule
    \multirow{5}[2]{*}{\dfaust} & nrICP & 79.78$\pm$109.09 & 4.09$\pm$9.38 & 74.79$\pm$14.96 & - \\
          & AN    & 31.74$\pm$42.13 & 0.90$\pm$2.81 & 91.88$\pm$5.49 & 0.34$\pm$0.06 \\
          & DSR   & 68.79$\pm$61.49 & 3.76$\pm$5.28 & 78.00$\pm$6.34 & 11.21$\pm$3.10 \\
          & CC    & 29.57$\pm$67.84 & 1.12$\pm$5.53 & 94.35$\pm$10.10 & - \\
          & \textbf{OUR} & \boldmath{}\textbf{18.38$\pm$16.86}\unboldmath{} & \boldmath{}\textbf{0.35$\pm$0.90}\unboldmath{} & \boldmath{}\textbf{96.75$\pm$1.59}\unboldmath{} & \boldmath{}\textbf{0.32$\pm$0.05}\unboldmath{} \\
    \bottomrule
    \end{tabular}%
	\endgroup
	}
	\label{tab:results_clean_data}%
\end{table}%

%%%%%%%%%%%%%%%%%%%%%%%%%%%%%%%%%%%%%%%%%%%%%%%%%%%%%%%%%%%%%%%%%%%%%%%%%%%%%%%%
\parag{Sequences without Global Rotations.}
In Table~\ref{tab:results_clean_data}, we report results on \anim{}, \amaa{} and \dfaust{}, none of which features significant global rotations. Our method systematically yields the most accurate correspondences. It also delivers the best reconstruction quality in Chamfer distance.

%%%%%%%%%%%%%%%%%%%%%%%%%%%%%%%%%%%%%%%%%%%%%%%%%%%%%%%%%%%%%%%%%%%%%%%%%%%%%%%%
\parag{Sequences with Global Rotations.}
In Table~\ref{tab:results_global_rotation}, we report results on \animr{} and \ama{} that feature rotating objects undergoing non-rigid deformations. Thanks to the loss term $\lossrot$, \ours{} generates patches that correctly follow the rotation of the object, which is reflected by much better values of the correspondence metrics $\mdist$, $\mrank$ and $\mpckauc$ than for the other methods.

\begin{table}[htbp]
  \centering
  \caption{\textbf{Comparison of \ours{} to SotA methods on correspondence accuracy and reconstruction quality using the datasets with global object rotation.} Our method is the most accurate and also yields reconstruction quality competitive with \atlasnet{}.}
  \resizebox{0.49\textwidth}{!}{
  	\begingroup
  	\setlength{\tabcolsep}{1pt}
  	\renewcommand{\arraystretch}{1.1}
    \begin{tabular}{clcccc}
    \toprule
    \textbf{dataset} & \textbf{\phantom{sp}model\phantom{sp}} & \boldmath{}\textbf{\phantom{spac}sL2 $\downarrow$\phantom{spac}}\unboldmath{} & \boldmath{}\textbf{\phantom{spa}rank $\downarrow$\phantom{spa}}\unboldmath{} & \boldmath{}\textbf{\phantom{sp}PCK auc $\uparrow$\phantom{sp}}\unboldmath{} & \boldmath{}\textbf{\phantom{spac}CD $\downarrow$\phantom{spac}}\unboldmath{} \\
    \midrule
    \multirow{6}[2]{*}{\animr} & nrICP & 187.26$\pm$190.35 & 20.87$\pm$26.11 & 42.78$\pm$29.30 & - \\
          & AN    & 188.88$\pm$182.15 & 21.32$\pm$25.40 & 43.15$\pm$30.23 & \boldmath{}\textbf{0.10$\pm$0.00}\unboldmath{} \\
          & DSR   & 218.05$\pm$165.69 & 25.28$\pm$23.79 & 30.16$\pm$23.70 & 12.46$\pm$3.05 \\
          & CC    & 98.55$\pm$127.10 & 11.25$\pm$15.92 & 72.45$\pm$23.45 & - \\
          & \textbf{OUR} & \boldmath{}\textbf{13.80$\pm$11.75}\unboldmath{} & \boldmath{}\textbf{0.39$\pm$0.67}\unboldmath{} & \boldmath{}\textbf{97.81$\pm$0.64}\unboldmath{} & 0.11$\pm$0.00 \\
    \midrule
    \multirow{6}[2]{*}{\ama} & nrICP & 224.10$\pm$189.77 & 12.70$\pm$15.15 & 36.50$\pm$24.55 & - \\
          & AN    & 223.23$\pm$200.53 & 14.03$\pm$16.71 & 45.41$\pm$22.78 & 0.29$\pm$0.01 \\
          & DSR   & 223.60$\pm$178.70 & 12.72$\pm$13.75 & 39.70$\pm$19.84 & 13.18$\pm$4.13 \\
          & CC    & 169.79$\pm$181.82 & 9.78$\pm$14.04 & 59.90$\pm$25.76 & - \\
          & \textbf{OUR} & \boldmath{}\textbf{49.59$\pm$46.46}\unboldmath{} & \boldmath{}\textbf{0.99$\pm$2.03}\unboldmath{} & \boldmath{}\textbf{84.04$\pm$9.97}\unboldmath{} & \boldmath{}\textbf{0.27$\pm$0.01}\unboldmath{} \\
    \bottomrule
    \end{tabular}%
  	\endgroup
	}
	\label{tab:results_global_rotation}%
\end{table}%

%%%%%%%%%%%%%%%%%%%%%%%%%%%%%%%%%%%%%%%%%%%%%%%%%%%%%%%%%%%%%%%%%%%%%%%%%%%%%%%
\parag{Sequences with Noisy Data.}\label{sec:noisy_data}
Table \ref{tab:results_noisy_data} summarizes our results on the noisy datasets \amaan{}, \cape{} and \inria{}. Comparing the results obtained on \amaan{} against those obtained on \amaa{}, the noiseless version of the dataset, and reported in Table~\ref{tab:results_clean_data}, reveals that the correspondence and reconstruction accuracies decrease slightly for all the methods. However, \ours{} still delivers the best correspondence accuracy by a substantial margin while still returning high fidelity reconstructions. Interestingly, \dsr{} yields the highest reconstruction accuracy as measured by CD on \amaan{} and \cape{} (while its training failed to converge on a some sequences of \cape{}). This is due to the strong distortion regularization, which keeps the individual patches rigid and thus robust against spatial noise. As before, \ours{} yields the highest correspondence accuracy on all the datasets.

\begin{table}[htbp]
  \centering
  \caption{\textbf{Comparison of \ours{} to SotA methods on correspondence accuracy and reconstruction quality using noisy data.} Our method is the most accurate in terms of predicted correspondences and also yields competitive reconstruction quality.}
  \resizebox{0.49\textwidth}{!}{
  	\begingroup
  	\setlength{\tabcolsep}{1pt}
  	\renewcommand{\arraystretch}{1.1}
    \begin{tabular}{clcccc}
    \toprule
    \textbf{dataset} & \textbf{\phantom{sp}model\phantom{sp}} & \boldmath{}\textbf{\phantom{spac}$\mdist$ $\downarrow$\phantom{spac}}\unboldmath{} & \boldmath{}\textbf{\phantom{spa}$\mrank$ $\downarrow$\phantom{spa}}\unboldmath{} & \boldmath{}\textbf{\phantom{sp}$\mpckauc$ $\uparrow$\phantom{sp}}\unboldmath{} & \boldmath{}\textbf{\phantom{spac}CD $\downarrow$\phantom{spac}}\unboldmath{} \\
    \midrule
    \multirow{5}[2]{*}{\amaan} & nrICP & 155.42$\pm$132.13 & 6.69$\pm$10.03 & 41.96$\pm$21.27 & - \\
          & AN    & 100.52$\pm$101.00 & 3.49$\pm$7.09 & 61.90$\pm$16.43 & 0.59$\pm$0.01 \\
          & DSR   & 115.53$\pm$130.46 & 5.24$\pm$10.34 & 61.85$\pm$18.65 & \boldmath{}\textbf{0.47$\pm$0.02}\unboldmath{} \\
          & CC    & 82.31$\pm$90.50 & 2.56$\pm$5.91 & 70.85$\pm$15.66 & - \\
          & \textbf{OUR} & \boldmath{}\textbf{46.32$\pm$42.85}\unboldmath{} & \boldmath{}\textbf{0.73$\pm$2.11}\unboldmath{} & \boldmath{}\textbf{84.22$\pm$7.75}\unboldmath{} & 0.58$\pm$0.02 \\
    \midrule
    \multirow{5}[2]{*}{\cape} & nrICP & 130.60$\pm$139.32 & 7.87$\pm$11.94 & 56.45$\pm$21.02 & - \\
          & AN    & 60.69$\pm$74.50 & 2.94$\pm$5.90 & 79.75$\pm$12.06 & \boldmath{}\textbf{0.35$\pm$0.06}\unboldmath{} \\
          & DSR   & 176.42$\pm$138.94 & 12.03$\pm$10.08 & 62.66$\pm$8.36 & 83.79$\pm$86.21 \\
          & CC    & 66.23$\pm$129.25 & 3.96$\pm$11.40 & 83.65$\pm$20.24 & - \\
          & \textbf{OUR} & \boldmath{}\textbf{26.24$\pm$34.87}\unboldmath{} & \boldmath{}\textbf{0.76$\pm$2.10}\unboldmath{} & \boldmath{}\textbf{93.84$\pm$4.90}\unboldmath{} & 0.42$\pm$1.76 \\
    \midrule
    \multirow{5}[2]{*}{\inria} & nrICP & 70.75$\pm$114.15 & 7.68$\pm$13.77 & 71.52$\pm$23.56 & - \\
          & AN    & 67.03$\pm$111.98 & 7.11$\pm$12.99 & 72.58$\pm$23.59 & 11.90$\pm$1.38 \\
          & DSR   & 83.64$\pm$121.01 & 9.49$\pm$15.06 & 67.95$\pm$25.06 & \boldmath{}\textbf{12.07$\pm$1.25}\unboldmath{} \\
          & CC    & 59.94$\pm$108.29 & 6.25$\pm$12.42 & 75.54$\pm$24.71 & - \\
          & \textbf{OUR} & \boldmath{}\textbf{26.80$\pm$46.58}\unboldmath{} & \boldmath{}\textbf{2.95$\pm$5.45}\unboldmath{} & \boldmath{}\textbf{89.27$\pm$13.48}\unboldmath{} & 12.12$\pm$1.59 \\
    \bottomrule
    \end{tabular}%
  	\endgroup
	}
	\label{tab:results_noisy_data}%
\end{table}%

%%%%%%%%%%%%%%%%%%%%%%%%%%%%%%%%%%%%%%%%%%%%%%%%%%%%%%%%%%%%%%%%%%%%%%%%%%%%%%%
\subsection{Ablation Study}
\label{sec:ablation}

We now subject our approach to a detailed analysis to demonstrate the importance of loss terms $\lossmc$ and $\lossrot$ and of a careful sampling strategy both in terms of data pair selection and progressive data sampling. For simplicity, we perform this ablation study on the clean datasets using only the corresponding validation sequences, that is, \texttt{cat\_walk} from \anim{}, \texttt{crane} from \ama{} and \texttt{jumping\_jacks} from \dfaust{}. However, as shown in Section \ref{sec:noisy_data}, our findings and hyperparameter selections also hold for the noisy data.

%===============================================================================
\parag{Sampling Strategy for Training Pairs.}
Instead of using time-adjacent point-cloud pairs (\textit{adjacent}), we can use random pairs (\textit{random}). We report the results in Table~\ref{tab:pairs_sampling_strategy}, where \textit{adjacent} clearly yields higher correspondence accuracy. In the particular case of dataset \ama{}, the model trained using \textit{random} produces mappings which fail to track the rotating object, hence the extremely high error.

\begin{table}[htbp]
  \centering
  \caption{\textbf{Switching pair sampling strategy} from adjacent, the default, to random.}
  \resizebox{0.49\textwidth}{!}{
  	\begingroup
  	\setlength{\tabcolsep}{1pt}
  	\renewcommand{\arraystretch}{1.1}
    \begin{tabular}{cccccc}
    \toprule
    \textbf{sequence} & \multicolumn{1}{l}{\textbf{\phantom{sp}strategy\phantom{sp}}} & \boldmath{}\textbf{\phantom{spac}$\mdist$ $\downarrow$\phantom{spac}}\unboldmath{} & \boldmath{}\textbf{\phantom{spa}$\mrank$ $\downarrow$\phantom{spa}}\unboldmath{} & \boldmath{}\textbf{\phantom{sp}$\mpckauc$ $\uparrow$\phantom{sp}}\unboldmath{} & \boldmath{}\textbf{\phantom{spac}CD $\downarrow$\phantom{spac}}\unboldmath{} \\
    \midrule
    \multirow{2}[2]{*}{\makecell{\anim\\(cat)}} & random & 54.48$\pm$84.92 & 3.92$\pm$8.53 & 81.07$\pm$12.62 & 0.446$\pm$0.007 \\
          & adjacent & \textbf{8.45$\pm$12.93} & \textbf{0.20$\pm$0.56} & \textbf{98.61$\pm$0.61} & \textbf{0.375$\pm$0.003} \\
    \midrule
    \multirow{2}[2]{*}{\makecell{\ama\\(crane)}} & random & 226.21$\pm$232.50 & 13.90$\pm$19.96 & 40.60$\pm$32.20 & 0.330$\pm$0.018 \\
          & adjacent & \textbf{46.92$\pm$35.52} & \textbf{0.55$\pm$1.03} & \textbf{83.27$\pm$8.87} & \textbf{0.280$\pm$0.005} \\
    \midrule
    \multirow{2}[2]{*}{\makecell{\dfaust\\(jacks)}} & random & 25.06$\pm$39.32 & 0.55$\pm$2.45 & 94.46$\pm$2.97 & 0.357$\pm$0.054 \\
          & adjacent & \textbf{23.96$\pm$33.16} & \textbf{0.48$\pm$1.91} & \textbf{94.78$\pm$2.54} & \textbf{0.339$\pm$0.046} \\
    \bottomrule
    \end{tabular}%
	\endgroup
	}
	\label{tab:pairs_sampling_strategy}%
\end{table}%

%===============================================================================
\parag{Influence of the Loss Term $\lossmc$.}
We change the value of the hyper-parameter $\alphamc$ of Eq.~\ref{eq:fullLoss} that balances metric consistency against Chamfer distance accuracy, with $\alpharot$  set to $0.1$. Setting $\alphamc$ too low turns off $\lossmc$ while setting it too high overpowers $\losscd$, which imposes strict isometry and makes the position of the patches ambiguous. As can be seen in Table~\ref{tab:strength_Lmc}, the choice of $\alphamc=0.1$ consistently yields the most accurate correspondences across all three datasets and thus we used it in all our experiments.

\begin{table}[htbp]
  \centering
  \caption{\textbf{The impact of the loss term $\lossmc$ on the accuracy of correspondences as measured by $\mdist$.}}
  \resizebox{0.49\textwidth}{!}{
  	\begingroup
  	\setlength{\tabcolsep}{1pt}
  	\renewcommand{\arraystretch}{1.0}
    \begin{tabular}{lccccc}
    \toprule
    \boldmath{}\textbf{$\alphamc$}\unboldmath{} & \boldmath{}\textbf{\phantom{spac}$1e^{-3}$\phantom{spac}}\unboldmath{} & \boldmath{}\textbf{\phantom{spac}$1e^{-2}$\phantom{spac}}\unboldmath{} & \textbf{\phantom{spac}0.1\phantom{spac}} & \textbf{\phantom{spac}1\phantom{spac}} & \boldmath{}\textbf{\phantom{spac}$1e^{1}$\phantom{spac}}\unboldmath{} \\
    \midrule
    \anim{} (cat) & 10.70$\pm$14.56 & 9.91$\pm$14.14 & \textbf{8.45$\pm$12.93} & 11.38$\pm$16.31 & 12.36$\pm$17.64 \\
    \ama{} (crane) & 56.62$\pm$38.79 & 62.04$\pm$48.79 & \textbf{46.92$\pm$35.52} & 102.17$\pm$151.91 & 253.14$\pm$252.75 \\
    \dfaust{} (jacks) & 28.03$\pm$31.53 & 24.18$\pm$34.46 & \textbf{23.96$\pm$33.16} & 28.31$\pm$40.78 & 30.43$\pm$54.09 \\
    \bottomrule
    \end{tabular}%
	\endgroup
	}
	\label{tab:strength_Lmc}%
\end{table}%	

%===============================================================================
\parag{Influence of the Loss Term $\lossrot$.}
We now vary the value of  $\alpharot$ of Eq.~\ref{eq:fullLoss}, while keeping $\alphamc$ fixed to 0.1. We report our results in Table~\ref{tab:strength_Lssc}. The results reveal that setting $\alpharot$ too high makes $\lossrot$ overpower the metric consistency term $\lossmc$ and consequently the model fails to recover semantically correct correspondences when the object rotates. Since choosing $\alpharot=0.1$ yields the best results, we kept it for all the experiments.

\begin{table}[htbp]
  \centering
	\caption{\textbf{The impact of the loss term $\lossrot$ on the accuracy of correspondences as measured by $\mdist$.}}
	\resizebox{0.49\textwidth}{!}{
	\begingroup
	\setlength{\tabcolsep}{1pt}
	\renewcommand{\arraystretch}{1.0}
    \begin{tabular}{lccccc}
    \toprule
    \boldmath{}\textbf{$\alpharot$\phantom{spacer}}\unboldmath{} & \boldmath{}\textbf{\phantom{spac}$1e^{-3}$\phantom{spac}}\unboldmath{} & \boldmath{}\textbf{\phantom{spac}$1e^{-2}$\phantom{spac}}\unboldmath{} & \textbf{\phantom{space}0.1\phantom{space}} & \textbf{\phantom{spac}1\phantom{space}} & \boldmath{}\textbf{\phantom{space}$1e^{1}$\phantom{spac}}\unboldmath{} \\
    \midrule
    \ama{} (crane) & 53.14$\pm$39.97 & 76.70$\pm$120.57 & \textbf{46.92$\pm$35.52} & 67.83$\pm$59.83 & 181.73$\pm$179 \\
    \bottomrule
    \end{tabular}%
	\endgroup
	}
	\label{tab:strength_Lssc}%
\end{table}%

%===============================================================================
\parag{Turning on and off Loss Terms.}\label{sec:turning_on_and_off_loss_terms}
In this experiment we turn on and off the individual loss terms. The ticks and crosses in Table~\ref{tab:ablation_study} indicate whether the given loss term is present, with its corresponding hyper-parameter set to $0.1$, or absent.  While adding either $\lossmc$ or $\lossrot$ to the vanilla model---similar to \atlasnet{}---improves the correspondence metrics, the highest accuracy is only reached when both terms are present.

\begin{table}[htbp]
  \centering
  \caption{\textbf{The impact of the individual loss terms on the reconstruction and correspondences accuracy.} The experiment was performed on the validation sequence \texttt{crane} from \ama which involves a global rotation of the object.}
	\resizebox{0.49\textwidth}{!}{
  	\begingroup
  	\setlength{\tabcolsep}{1pt}
  	\renewcommand{\arraystretch}{1.1}
    \begin{tabular}{cccccc}
    \toprule
    \boldmath{}\textbf{\phantom{s}$\alphamc$\phantom{s}}\unboldmath{} & \boldmath{}\textbf{\phantom{s}$\alpharot$\phantom{s}}\unboldmath{} & \boldmath{}\textbf{\phantom{space}$\mdist$ $\downarrow$\phantom{space}}\unboldmath{} & \boldmath{}\textbf{\phantom{spac}$\mrank$ $\downarrow$\phantom{spac}}\unboldmath{} & \boldmath{}\textbf{\phantom{spa}$\mpckauc$ $\uparrow$\phantom{spa}}\unboldmath{} & \phantom{space}CD $\downarrow$\phantom{space} \\
    \midrule
    \xmark & \xmark & 317.02$\pm$293.69 & 21.78$\pm$25.99 & 29.70$\pm$30.18 & 0.283$\pm$0.010 \\
    \cmark & \xmark & 247.44$\pm$253.65 & 15.14$\pm$21.38 & 36.86$\pm$34.17 & \textbf{0.270$\pm$0.010} \\
    \xmark & \cmark & 95.47$\pm$92.53 & 2.63$\pm$4.99 & 59.30$\pm$22.24 & 0.272$\pm$0.005 \\
    \cmark & \cmark & \textbf{46.92$\pm$35.52} & \textbf{0.55$\pm$1.03} & \textbf{83.27$\pm$8.87} & 0.280$\pm$0.005 \\
    \bottomrule
    \end{tabular}%
  	\endgroup
	}
	\label{tab:ablation_study}%
\end{table}%

%===============================================================================
\parag{Progressive Data Sampling.}
Here, we compare the \textit{progressive} sampling of Section \ref{sec:progressive_data_sampling} to directly sampling the \textit{whole} 
%section. 
sequence. As can be seen from Table~\ref{tab:progressive_ds_sampling}, while the reconstruction quality as measured by \mcd{} is affected only marginally, the correspondence accuracy significantly degrades when switching from \textit{progressive} to \textit{whole}. This is caused by the strong symmetries present in the data, which can result in swaps between the person's front and back. This does not affect the reconstruction quality but degrades the correspondences. 

\begin{table}[htbp]
  \centering
  \caption{\textbf{The impact of the progressive sampling strategy on the reconstruction and correspondences accuracy.} The experiments were run on the validation sequence \texttt{crane} from \ama{} which involves a global rotation of the object.}
  	\resizebox{0.49\textwidth}{!}{
  	\begingroup
  	\setlength{\tabcolsep}{1pt}
  	\renewcommand{\arraystretch}{1.1}
    \begin{tabular}{ccccc}
    \toprule
    \multicolumn{1}{c}{\textbf{strategy}} & \boldmath{}\textbf{\phantom{spacer}$\mdist$ $\downarrow$\phantom{spacer}}\unboldmath{} & \boldmath{}\textbf{\phantom{spac}$\mrank$ $\downarrow$\phantom{spac}}\unboldmath{} & \boldmath{}\textbf{\phantom{spa}$\mpckauc$ $\uparrow$\phantom{spa}}\unboldmath{} & \phantom{spacer}CD $\downarrow$\phantom{spacer} \\
    \midrule
    normal & 268.97$\pm$250.53 & 17.71$\pm$21.77 & 34.53$\pm$34.44 & \textbf{0.278$\pm$0.010} \\
    progressive & \textbf{46.92$\pm$35.52} & \textbf{0.55$\pm$1.03} & \textbf{83.27$\pm$8.87} & 0.280$\pm$0.005 \\
    \bottomrule
    \end{tabular}%
	\endgroup
	}
	\label{tab:progressive_ds_sampling}%
\end{table}%

%===============================================================================
\parag{Impact of $\lossrot$ on Data without Global Rotation.} \label{sec:impact_of_lossrot_on_data_without_global_rotation}
When reconstructing pre-aligned, neither the loss term $\lossrot$ nor the progressive data sampling strategy is theoretically needed. The Table~\ref{tab:impact_of_lossrot_on_data_without_glob_rot} show that if we remove both (\oursnolrot{}), indeed, the accuracy of the predicted correspondences does not improve or degrade overall.

\begin{table}[htbp]
  \centering
  \caption{\textbf{The impact of $\lossrot$ together with progressive sampling strategy on the reconstruction and correspondences accuracy of the clean aligned data.} The results represent the average over all the sequences, except the validation ones, of the given dataset.}
  \resizebox{0.49\textwidth}{!}{
  	\begingroup
  	\setlength{\tabcolsep}{1pt}
  	\renewcommand{\arraystretch}{1.1}
    \begin{tabular}{clcccc}
    \toprule
    \textbf{dataset} & \textbf{\phantom{sp}model\phantom{sp}} & \boldmath{}\textbf{\phantom{spac}$\mdist$ $\downarrow$\phantom{spac}}\unboldmath{} & \boldmath{}\textbf{\phantom{spa}$\mrank$ $\downarrow$\phantom{spa}}\unboldmath{} & \boldmath{}\textbf{\phantom{sp}$\mpckauc$ $\uparrow$\phantom{sp}}\unboldmath{} & \boldmath{}\textbf{\phantom{spac}CD $\downarrow$\phantom{spac}}\unboldmath{} \\
    \midrule
    \multirow{2}[2]{*}{\anim} & OUR w/o $\lossrot$ & \boldmath{}\textbf{11.93$\pm$11.00}\unboldmath{} & \boldmath{}\textbf{0.30$\pm$0.57}\unboldmath{} & \boldmath{}\textbf{98.10$\pm$0.61}\unboldmath{} & \boldmath{}\textbf{0.09$\pm$0.00}\unboldmath{} \\
          & \textbf{OUR} & 12.60$\pm$11.38 & 0.36$\pm$0.72 & 98.00$\pm$0.71 & \boldmath{}\textbf{0.09$\pm$0.00}\unboldmath{} \\
    \midrule
    \multirow{2}[2]{*}{\amaa} & OUR w/o $\lossrot$ & 57.12$\pm$65.33 & 1.55$\pm$3.90 & 82.29$\pm$11.16 & 0.32$\pm$0.02 \\
          & \textbf{OUR} & \boldmath{}\textbf{35.60$\pm$29.42}\unboldmath{} & \boldmath{}\textbf{0.36$\pm$0.94}\unboldmath{} & \boldmath{}\textbf{89.73$\pm$5.26}\unboldmath{} & \boldmath{}\textbf{0.27$\pm$0.01}\unboldmath{} \\
    \midrule
    \multirow{2}[2]{*}{\dfaust} & OUR w/o $\lossrot$ & 19.81$\pm$20.96 & 0.38$\pm$1.07 & 96.17$\pm$2.24 & 0.34$\pm$0.06 \\
          & \textbf{OUR} & \boldmath{}\textbf{18.38$\pm$16.86}\unboldmath{} & \boldmath{}\textbf{0.35$\pm$0.90}\unboldmath{} & \boldmath{}\textbf{96.75$\pm$1.59}\unboldmath{} & \boldmath{}\textbf{0.32$\pm$0.05}\unboldmath{} \\
    \bottomrule
    \end{tabular}%
	\endgroup
	}
	\label{tab:impact_of_lossrot_on_data_without_glob_rot}%
\end{table}%

%===============================================================================
\parag{Impact of $\lossrot$ on Data with Noise.} \label{sec:impact_of_lossrot_on_data_with_noise}
On the contrary, in case of the noisy data, the combination of the loss function $\lossrot$ and the progressive dataset sampling consistently improves the correspondences accuracy, as shown in Table~\ref{tab:impact_of_lossrot_on_data_with_noise}. This can be attributed to the fact that augmenting the data by generating random rotations of the deforming object, as explained in Section \ref{sec:rotations}, regularizes the learned mappings effectively making the model more robust against the noise.

\begin{table}[htbp]
  \centering
  \caption{\textbf{The impact of $\lossrot$ together with progressive sampling strategy on the reconstruction and correspondences accuracy of the data with noise.} The results represent the average over all the sequences, except the validation ones, of the given dataset.}
  \resizebox{0.49\textwidth}{!}{
  	\begingroup
  	\setlength{\tabcolsep}{1pt}
  	\renewcommand{\arraystretch}{1.1}
    \begin{tabular}{clcccc}
    \toprule
    \textbf{dataset} & \textbf{\phantom{sp}model\phantom{sp}} & \boldmath{}\textbf{\phantom{spac}$\mdist$ $\downarrow$\phantom{spac}}\unboldmath{} & \boldmath{}\textbf{\phantom{spa}$\mrank$ $\downarrow$\phantom{spa}}\unboldmath{} & \boldmath{}\textbf{\phantom{sp}$\mpckauc$ $\uparrow$\phantom{sp}}\unboldmath{} & \boldmath{}\textbf{\phantom{spac}CD $\downarrow$\phantom{spac}}\unboldmath{} \\
    \midrule
    \multirow{2}[2]{*}{\amaan} & OUR w/o $\lossrot$ & 69.78$\pm$84.98 & 2.15$\pm$5.28 & 77.79$\pm$12.85 & \boldmath{}\textbf{0.58$\pm$0.01}\unboldmath{} \\
          & \textbf{OUR} & \boldmath{}\textbf{46.32$\pm$42.85}\unboldmath{} & \boldmath{}\textbf{0.73$\pm$2.11}\unboldmath{} & \boldmath{}\textbf{84.22$\pm$7.75}\unboldmath{} & \boldmath{}\textbf{0.58$\pm$0.02}\unboldmath{} \\
    \midrule
    \multirow{2}[2]{*}{\cape} & OUR w/o $\lossrot$ & 28.01$\pm$37.97 & 0.81$\pm$2.22 & 93.05$\pm$5.47 & 0.47$\pm$1.96 \\
          & \textbf{OUR} & \boldmath{}\textbf{26.24$\pm$34.87}\unboldmath{} & \boldmath{}\textbf{0.76$\pm$2.10}\unboldmath{} & \boldmath{}\textbf{93.84$\pm$4.90}\unboldmath{} & \boldmath{}\textbf{0.42$\pm$1.76}\unboldmath{} \\
    \midrule
    \multirow{2}[2]{*}{\inria} & OUR w/o $\lossrot$ & 61.77$\pm$111.54 & 6.77$\pm$13.76 & 75.89$\pm$27.12 & \boldmath{}\textbf{12.00$\pm$1.28}\unboldmath{} \\
          & \textbf{OUR} & \boldmath{}\textbf{26.80$\pm$46.58}\unboldmath{} & \boldmath{}\textbf{2.95$\pm$5.45}\unboldmath{} & \boldmath{}\textbf{89.27$\pm$13.48}\unboldmath{} & 12.12$\pm$1.59 \\
    \bottomrule
    \end{tabular}%
	\endgroup
	}
	\label{tab:impact_of_lossrot_on_data_with_noise}%
\end{table}%

%% file: sections/conclusion.tex
\section{Conclusion}
We have introduced an atlas-based method that yields temporally-consistent surface reconstructions in an unsupervised manner, by enforcing a point on the canonical shape representation to map to metrically-consistent 3D points on the reconstructed surfaces.

While our method yields  better surface correspondences than  state-of-the-art surface reconstruction techniques, it shares one shortcoming with these atlas-based methods: The reconstructed patches may overlap, causing imperfections in the reconstructions. Another limitation of our method is that we use heuristics for the hyper-parameters balancing metric-consistency, self-supervised correspondences and reconstruction; employing an annealing-like technique which gradually permits more non-isometric deformations may be the next logical step. Furthermore, our framework could easily extend to other distortion measures, such as a conformal one. Studying this for non-isometric reconstruction and matching would be an interesting direction for future work. 
Finally, we envision many future applications for our  approach. For example, replacing the Chamfer distance with an image-based loss would allow us to apply our method to RGB video sequences, which we believe could instigate progress in video-based 3D reconstruction.

%% file: sections/acknowledgments.tex
\section*{Acknowledgments}
This work was partially carried out while the first author was an intern at Adobe Research and was funded in part by the Swiss National Science Foundation.

%% file: sections/supplementary.tex
\appendices

%%%%%%%%%%%%%%%%%%%%%%%%%%%%%%%%%%%%%%%%%%%%%%%%%%%%%%%%%%%%%%%%%%%%%%%%%%%%%%%%%%%%%%%%%%%%%%%%%%%%

\section{Training and Evaluation Details}
We provide details of the triplet sampling strategy used to train the cycle consistent point cloud deformation method~\cite{Groueix19} (\cyccon{}) in Section~\ref{ssec:details_of_trianing_cc}, an analysis of the strategy used to evaluate the non-rigid ICP method~\cite{Huang17f} (\nricp{}) in Section~\ref{ssec:analysis_of_nricp} and more information on the points sampling strategy used to evaluate all the atlas-based methods, i.e. AtlasNet~\cite{Groueix18b} (\atlasnet{}), Differential Surface Representation~\cite{Bednarik20a} (\dsr{}) and our method (\ours{}), in Section~\ref{ssec:point_sampling_in_atlas_based_methods}.

%---------------------------------------------------------------------------------------------------
\subsection{Details of Training \cyccon{}} \label{ssec:details_of_trianing_cc}

The training of \cyccon{} relies on sampling triplets of shapes from the given dataset. The authors argue that the best results were achieved when sampling triplets of shapes that are close to each other in the Chamfer distance (CD) sense. Specifically, given a randomly sampled shape $A$, two other shapes $B, C$ are randomly sampled from the $20$ nearest neighbors of $A$ to complete the triplet. Let us refer to this sampling strategy as \textit{knn}.

\ours{} itself relies on sampling shape pairs, and as shown in Section 4.3 of the main paper, better results are achieved when sampling the shape pairs from a time window $\delta$ of a given sequence (\textit{neighbors}) rather sampling pairs randomly within a sequence (\textit{random}).

For fair comparison, we experimented with training \cyccon{} using all three strategies, \textit{knn}, \textit{neighbors} and \textit{random}. Table~\ref{tab:cc_triplet_sampling} reports the results on the DFAUST dataset using the validation sequence \texttt{jumping\_jacks} and one more randomly chosen sequence \texttt{jiggle\_on\_toes}. Since \cyccon{} performs best by a large margin when trained using \textit{random}, we use this strategy for all the experiments.

\begin{table}[htbp]
  \centering
  \caption{\textbf{Comparison of different triplet sampling strategies to train \cyccon{}}. The experiments were conducted using DFAUST.}
  \vspace{0.1cm}
    \resizebox{0.49\textwidth}{!}{
  	\begingroup
  	\setlength{\tabcolsep}{1pt}
  	\renewcommand{\arraystretch}{1.1}
    \begin{tabular}{ccccc}
    \toprule
    \textbf{sequence} & \phantom{spa}\textbf{sampling}\phantom{spa} & \boldmath{}\textbf{\phantom{spacer}$\mdist$ $\downarrow$\phantom{spacer}}\unboldmath{} & \boldmath{}\textbf{\phantom{spac}$\mrank$ $\downarrow$\phantom{spac}}\unboldmath{} & \boldmath{}\textbf{\phantom{spa} $\mpckauc$ $\uparrow$\phantom{spa}}\unboldmath{} \\
    \midrule
    \multirow{3}[2]{*}{jumping\_jacks} & knn   & 105.57$\pm$217.32 & 6.43$\pm$18.26 & 77.06$\pm$20.18 \\
          & neighbors & 95.13$\pm$179.81 & 6.21$\pm$17.42 & 75.84$\pm$20.78 \\
          & random & \boldmath{}\textbf{32.74$\pm$31.65}\unboldmath{} & \boldmath{}\textbf{0.70$\pm$1.68}\unboldmath{} & \boldmath{}\textbf{91.47$\pm$6.25}\unboldmath{} \\
    \midrule
    \multirow{3}[2]{*}{jiggle\_on\_toes} & knn   & 71.73$\pm$203.46 & 4.36$\pm$16.88 & 87.77$\pm$20.92 \\
          & neighbors & 47.69$\pm$99.55 & 2.15$\pm$8.48 & 88.31$\pm$12.20 \\
          & random & \boldmath{}\textbf{26.26$\pm$69.02}\unboldmath{} & \boldmath{}\textbf{0.91$\pm$5.71}\unboldmath{} & \boldmath{}\textbf{94.86$\pm$10.98}\unboldmath{} \\
    \bottomrule
    \end{tabular}%
  \endgroup
	}
  \label{tab:cc_triplet_sampling}%
\end{table}%

%---------------------------------------------------------------------------------------------------
\subsection{Analysis of the \nricp{} \cite{Huang17f}  Strategy} \label{ssec:analysis_of_nricp}
The \nricp{} method deforms a point cloud to best match another point cloud and thus can be used to find point-wise correspondences in an unsupervised way. Formally, let $\nu_{P_{j}}$ be the non-rigid ICP function which deforms an input point cloud $P_{i}$ to best match $P_{j}$. Following the notation introduced in Section 4.1 of the main paper, let $\pi_{\mathcal{X}}$ be a mapping that projects the points from an input point cloud to their respective nearest neighbors in the target point cloud $\mathcal{X}$. The simplest way to use \nricp{} to find correspondences between a pair of point clouds $(P_{i}, P_{j})$ randomly drawn from the given sequence is to compute $\pi_{P_{j}} \circ \nu_{P_{j}}(P_{i})$. Let us call this strategy \textit{random}.

Non-rigid ICP tends to break when the deformation between the two point clouds is severe. However, as we are dealing with sequences depicting a deforming shape, one can compute the correspondences between a pair of point clouds $(P_{i}, P_{j})$ by first predicting the correspondences for consecutive pairs of point clouds where the deformation is minimal, i.e., $(P_{i}, P_{i+1}), (P_{i+1}, P_{i+1}), \dots, (P_{j-1}, P_{j})$, and finally propagating the correspondences from $P_{i}$ to $P_{j}$. Formally, we compute $\pi_{P_{j}} \circ \nu_{P_{j}} \circ \dots \circ \pi_{P_{i+2}} \circ \nu_{P_{i+2}} \circ \pi_{P_{i+1}} \circ \nu_{P_{i+1}}(P_{i})$ and refer to this strategy as \textit{propagate\_simple}.

The drawback of \textit{propagate\_simple} is that every mapping $\pi_{P_{k}}$ is onto and thus throughout the propagation, progressively more source points get mapped to the same target point, which causes a loss of spatial information and ultimately yields less precise correspondences. To overcome this problem, one can replace $\pi_{P_{k}}$ with $\rho_{P_{k}}$, which performs a Hungarian matching of the input point cloud and the target point cloud $P_{k}$ with the objective of minimizing the overall per-point-pair distance. Formally, we compute $\rho_{P_{j}} \circ \nu_{P_{j}} \circ \dots \circ \rho_{P_{i+2}} \circ \nu_{P_{i+2}} \circ \rho_{P_{i+1}} \circ \nu_{P_{i+1}}(P_{i})$ and call this strategy \textit{propagate\_bijective}.

Finally, an alternative option is not to perform any projection $\pi_{P_{k}}$ or $\rho_{P_{k}}$ as we propagate the correspondences from $P_{i}$ to $P_{j}$, but instead to gradually deform the input point cloud $P_{i}$ to best match each point cloud along the sequence between $P_{i}$ and $P_{j}$. Formally, we compute $\nu_{P_{j}} \circ \dots \nu_{P_{i+2}} \circ \nu_{P_{i+1}}(P_{i})$ and refer to this strategy as \textit{propagate\_deform}.

Table \ref{tab:nricp_eval_strategy} reports the results of all four aforementioned correspondence estimation strategies on the \texttt{crane} validation sequence  from the AMA dataset. We found that \textit{propagate\_simple} suffers from the loss of spatial precision due to the onto mapping. While \textit{propagate\_bijective} overcomes this problem, the Hungarian matching introduces a strong drift along the sequence yielding even worse overall correspondences. The strategy \textit{propagate\_deform} performs the best out of all three propagation-based strategies, but is still outperformed by the simplest strategy \textit{random}. Therefore, as \textit{random} yields the highest correspondence accuracy, we use it to evaluate \nricp{} on all datasets.

\begin{table}[htbp]
  \centering
  \caption{\textbf{Comparison of strategies used to establish correspondences with \nricp{}}. The experiments were conducted on the \texttt{crane} validation sequence from the AMA dataset.}
  \vspace{0.1cm}
    \resizebox{0.49\textwidth}{!}{
  	\begingroup
  	\setlength{\tabcolsep}{1pt}
  	\renewcommand{\arraystretch}{1.1}
    \begin{tabular}{cccc}
    \toprule
    \textbf{strategy} & \boldmath{}\textbf{\phantom{spacer}$\mdist$ $\downarrow$\phantom{spacer}}\unboldmath{} & \boldmath{}\textbf{\phantom{space}$\mrank$ $\downarrow$\phantom{space}}\unboldmath{} & \boldmath{}\textbf{\phantom{space}$\mpckauc$ $\uparrow$\phantom{space}}\unboldmath{} \\
    \midrule
    random & \boldmath{}\textbf{172.55$\pm$167.76}\unboldmath{} & \boldmath{}\textbf{7.83$\pm$12.56}\unboldmath{} & \boldmath{}\textbf{41.61$\pm$19.29}\unboldmath{} \\
    propagate\_simple & 211.11$\pm$147.43 & 9.38$\pm$10.32 & 23.01$\pm$18.31 \\
    propagate\_bijective & 213.87$\pm$169.00 & 10.55$\pm$13.99 & 25.31$\pm$17.80 \\
    propagate\_deform & 206.64$\pm$150.45 & 10.40$\pm$13.35 & 25.41$\pm$16.55 \\
    \bottomrule
    \end{tabular}%
  \endgroup
	}
  \label{tab:nricp_eval_strategy}%
\end{table}%

%---------------------------------------------------------------------------------------------------
\subsection{Point Sampling in Atlas Based Methods} 
\label{ssec:point_sampling_in_atlas_based_methods}

The original AtlasNet work \cite{Groueix18b} argues that better reconstruction accuracy is achieved if the 2D points sampled from the UV domain $\uvdom$ are spaced on a regular grid. As explained in Section 4.1 of the main paper, at evaluation time each atlas based method, i.e., \atlasnet{}, \dsr{} and \ours{}, predicts $N = 3125$ points. Due to the unknown number of collapsed patches, which are discarded at runtime, it might not be possible to evenly split $N$ points into $P$ non-collapsed patches so that the points would form a regular grid in the UV space $\uvdom$.

\begin{figure}[tbh]
\begin{center}
\includegraphics[width=0.9\linewidth]{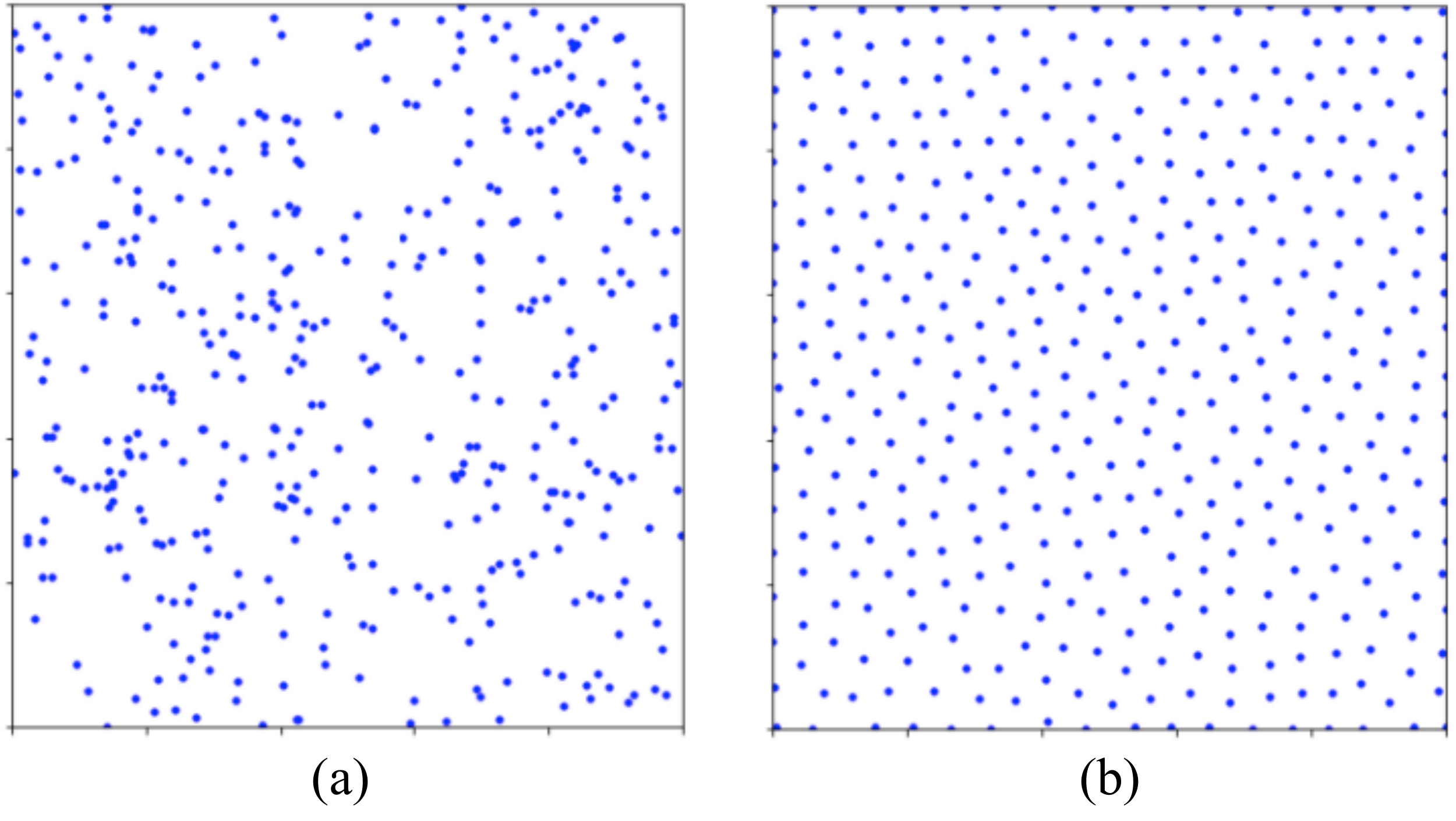}
\end{center}
  \caption{\textbf{Comparison of (a) uniform and (b) as regular as possible 2D points sampling.}}
\label{fig:regular_2d_pts}
\end{figure}

Therefore, instead of using a regular grid, we distribute the given available number of points as regularly as possible in the 2D domain using a simulated annealing based algorithm. The points are initially distributed uniformly at random, and then their position is iteratively adjusted so that every point maximizes its distance to the nearest points. This procedure is summarized in Algorithm \ref{alg:regular_2d_pts}. The difference between random and as regular as possible 2D points sampling is demonstrated in Fig. \ref{fig:regular_2d_pts}.

\begin{algorithm}[h]
\label{alg:regular_2d_pts}
\DontPrintSemicolon
  
  \KwInput{$M \in \mathbb{N}$  \tcp*{Number of 2D points.}}
  \KwOutput{$p_{i} \in \real^{2}, \forall 1\leq i \leq M$ \tcp*{2D points.}}
  \tcc{Initialization}
  step := $\frac{1}{4\sqrt{M}}$\;
  decay := 0.994\;
  $p_{i} \sim \mathcal{U}(\mathbf{0}, \mathbf{1}), \forall 1 \leq i \leq M$\tcp*{Random 2D points.}
  iter := 0\;
  \tcc{Main algorithm.}
  \While{$\text{iter} < 250$}
  {
        \For{$i:=1$ \KwTo $M$}{
      		$d_{i} :=\ \min_{j \neq i}{||p_{i} - p_{j}||}$\;
      		$\alpha_{i} \sim \mathcal{U}(0, 2\pi)$\;
      		$p_{i}^{\text{new}}$ := $p_{i} + \text{step} \cdot R(\alpha_{i})\begin{bmatrix}1 \\ 0\end{bmatrix}$\tcp*{R: rot. matrix} \; 
      		$d_{i}^{\text{new}} := \min_{j \neq i}{||p_{i}^{\text{new}} - p_{j}^{\text{new}}||}$\;
      		\If{$d_{i}^{\text{new}} > d_{i}$}
      		{
      		    $p_{i}$ := $p_{i}^{\text{new}}$
      		}
      	}
      	step := step $\cdot$ decay\;
      	iter := iter $+ 1$\;
  }
\caption{As regular as possible 2D points.}
\end{algorithm}

%---------------------------------------------------------------------------------------------------
\subsection{Time Complexity} \label{ssec:time_complexity}
The optimization of all the learning based methods was performed using an Nvidia Tesla V100 GPU, and processing a sequence of average length takes $23.0$ hours for OUR, while AN, DSR and CC take $4.1$, $16.4$ and $9.7$ hours, respectively. Note, that if it is known a-priory that the input sequence does not contain substantial global rotation, the loss term $\lossrot$ (and the progressive sampling strategy) can be turned off, in which case the optimization takes $16.1$ hours. nrICP does not involve the optimization stage and can process $\sim1$ sample per second.

%%%%%%%%%%%%%%%%%%%%%%%%%%%%%%%%%%%%%%%%%%%%%%%%%%%%%%%%%%%%%%%%%%%%%%%%%%%%%%%%%%%%%%%%%%%%%%%%%%%%
\section{Complete Results}

We provide details of the search for the best value of our method's hyperparameter $\delta$ in Section \ref{ssec:tuning_time_window} and we list the complete per-sequence results of all the evaluated methods on all the datasets in Section \ref{ssec:evaluation_on_all}. Furthermore, we refer the reader to the \href{https://youtu.be/P4imXONmtto}{supplementary video}\footnote{https://youtu.be/P4imXONmtto} which contains the comparison of all methods on multiple sequences from all the datasets.

%---------------------------------------------------------------------------------------------------
\subsection{Tuning the Time-Window $\delta$}\label{ssec:tuning_time_window}
As described in Section 3.3 of the main paper, \ours{} relies on sampling pairs of shapes from a time window denoted as $\delta$. We tuned this hyperparamater individually for different dataset using a respective validation sequence, and set it to the values yielding the best correspondence accuracy as measured by the metrics $\mdist, \mrank$ and $\mpckauc$. Similarly to the ablation study reported in Section~4.5 of the main paper, for simplicity we only consider the clean datasets \anim{}, \ama{} and \dfaust{} and use the findings for the remaining ones. Table~\ref{tab:search_delta} lists the results of training \ours{} for $\delta \in [1, 6]$ and justifies the selection of $\delta = 1$ for \anim{}, $\delta = 1$ for \ama{} and $\delta = 5$ for \dfaust{}.

Note that, as the \anim{} and \ama{} datasets have lower frame-rates than the \dfaust{} dataset, i.e., the surface undergoes larger motion from frame to frame, the correspondence error clearly decreases with the decreasing size of the time window $\delta$, indicating that our method benefits from observing pairs of shapes as similar as possible to each other. On the other hand, as the \dfaust{} dataset in general exhibits small frame to frame changes, the search reveals that our method can benefit from observing pairs from larger time windows. 

Since the frame-rate of the dataset \cape{} corresponds to that of \dfaust{} and the frame rate of the datasets \inria{} and \cmu{} correspond to that od \ama{}, we set $\delta = 5$ for \cape{} and $\delta = 1$ for \inria{} and \cmu{}.

\begin{table}[htbp]
  \centering
  \caption{\textbf{Search for the best value of the hyperparameter $\delta$ used by \ours{} on each dataset}.}
  \vspace{0.1cm}
  \resizebox{0.49\textwidth}{!}{
  	\begingroup
  	\setlength{\tabcolsep}{1pt}
  	\renewcommand{\arraystretch}{1.1}
    \begin{tabular}{cccccc}
    \toprule
    \textbf{dataset} & \textbf{neighbours} & \boldmath{}\textbf{\phantom{spac}$\mdist$ $\downarrow$\phantom{spac}}\unboldmath{} & \boldmath{}\textbf{\phantom{spa}$\mrank$ $\downarrow$\phantom{spa}}\unboldmath{} & \boldmath{}\textbf{\phantom{sp}$\mpckauc$ $\uparrow$\phantom{sp}}\unboldmath{} & \boldmath{}\textbf{\phantom{spac}CD $\downarrow$\phantom{spac}}\unboldmath{} \\
    \midrule
    \multirow{6}[2]{*}{\makecell{ANIM\\(cat)}} & 1     & \boldmath{}\textbf{8.45$\pm$12.93}\unboldmath{} & \boldmath{}\textbf{0.20$\pm$0.56}\unboldmath{} & \boldmath{}\textbf{98.61$\pm$0.61}\unboldmath{} & \boldmath{}\textbf{0.375$\pm$0.003}\unboldmath{} \\
          & 2     & 10.77$\pm$16.31 & 0.26$\pm$0.71 & 97.89$\pm$1.12 & 0.383$\pm$0.004 \\
          & 3     & 10.96$\pm$16.14 & 0.26$\pm$0.76 & 97.95$\pm$1.06 & 0.398$\pm$0.004 \\
          & 4     & 34.58$\pm$74.16 & 2.31$\pm$7.08 & 88.52$\pm$9.09 & 0.400$\pm$0.005 \\
          & 5     & 12.07$\pm$16.43 & 0.31$\pm$0.80 & 97.75$\pm$1.11 & 0.402$\pm$0.004 \\
          & 6     & 11.39$\pm$16.73 & 0.28$\pm$0.64 & 97.72$\pm$1.15 & 0.383$\pm$0.003 \\
    \midrule
    \multirow{6}[2]{*}{\makecell{AMA\\(crane)}} & 1     & \boldmath{}\textbf{46.92$\pm$35.52}\unboldmath{} & \boldmath{}\textbf{0.55$\pm$1.03}\unboldmath{} & \boldmath{}\textbf{83.27$\pm$8.87}\unboldmath{} & \boldmath{}\textbf{0.280$\pm$0.005}\unboldmath{} \\
          & 2     & 161.75$\pm$181.84 & 7.87$\pm$12.61 & 53.57$\pm$32.77 & 0.290$\pm$0.011 \\
          & 3     & 138.97$\pm$212.25 & 8.14$\pm$17.35 & 67.35$\pm$38.02 & 0.294$\pm$0.016 \\
          & 4     & 222.39$\pm$241.65 & 13.99$\pm$20.56 & 41.17$\pm$36.72 & 0.332$\pm$0.043 \\
          & 5     & 286.09$\pm$268.34 & 19.52$\pm$24.04 & 33.26$\pm$33.00 & 0.372$\pm$0.039 \\
          & 6     & 176.25$\pm$220.33 & 9.92$\pm$17.27 & 55.71$\pm$30.33 & 0.287$\pm$0.007 \\
    \midrule
    \multirow{6}[2]{*}{\makecell{DFAUST\\(jacks)}} & 1     & 25.98$\pm$39.40 & 0.66$\pm$2.71 & 94.47$\pm$2.83 & 0.364$\pm$0.047 \\
          & 2     & 25.81$\pm$36.08 & 0.57$\pm$2.54 & 94.51$\pm$2.83 & 0.377$\pm$0.049 \\
          & 3     & 24.54$\pm$34.38 & 0.57$\pm$2.15 & 95.17$\pm$2.29 & 0.365$\pm$0.047 \\
          & 4     & 24.73$\pm$35.96 & 0.51$\pm$2.19 & 94.77$\pm$2.53 & 0.361$\pm$0.049 \\
          & 5     & \boldmath{}\textbf{23.96$\pm$33.16}\unboldmath{} & \boldmath{}\textbf{0.48$\pm$1.91}\unboldmath{} & \boldmath{}\textbf{94.78$\pm$2.54}\unboldmath{} & \boldmath{}\textbf{0.339$\pm$0.046}\unboldmath{} \\
          & 6     & 27.33$\pm$38.36 & 0.65$\pm$2.49 & 93.85$\pm$3.67 & 0.370$\pm$0.053 \\
    \bottomrule
    \end{tabular}%
	\endgroup
	}
	\label{tab:search_delta}%
\end{table}%

%---------------------------------------------------------------------------------------------------
\subsection{Evaluation on all Datasets and Stress Test} \label{ssec:evaluation_on_all}

For brevity, Section~4.4.2 of the main paper only reports the mean results computed over all the sequences contained in the individual datasets. Here we report detailed results for each sequence separately. The results for the clean datasets \anim{}, \amaa{} and \dfaust{} are summarized in Tables \ref{tab:results_individual_anim}, \ref{tab:results_individual_amaa} and \ref{tab:results_individual_dfaust}, for the datasets with global rotations \animr{} and \ama{} in the Tables \ref{tab:results_individual_animr} and \ref{tab:results_individual_ama} and for the noisy datasets in Tables \ref{tab:results_individual_amaan}, \ref{tab:results_individual_cape} and \ref{tab:results_individual_inria}. Note that the average values reported in the last cell in each table are computed on all the test sequences, i.e., excluding the validation sequence \texttt{cat\_walk} in \anim{} and \animr{}, \texttt{crane} in \ama{}, \amaa{} and \amaan{}, \texttt{jumping\_jacks} in \dfaust{}.

Finally, Table \ref{tab:results_individual_stress_test} shows the results on the \texttt{horse\_collapse} sequence used for the stress test of our method, as shown in Fig.~11 of the main paper, and an additional similar sequence \texttt{camel\_collapse}. Both  sequences come from the same work of \cite{Sumner04} as the sequences \texttt{horse\_gallop}, \texttt{camel\_gallop} and \texttt{elephant\_gallop} from the \anim{} dataset, and thus we preprocess them in the same way, i.e., by scaling each sample so that the first frame of each sequence fits in a unit cube.

\begin{table}[htbp]
  \centering
  \caption{\textbf{Comparison of \ours{} to SotA methods on correspondence accuracy and reconstruction quality on the \anim{} dataset.}}
  \vspace{0.1cm}
  \resizebox{0.49\textwidth}{!}{
  	\begingroup
  	\setlength{\tabcolsep}{1pt}
  	\renewcommand{\arraystretch}{1.1}
\begin{tabular}{clcccc}
    \toprule
    \textbf{\phantom{spac}sequence\phantom{spac}} & \textbf{model} & \boldmath{}\textbf{\phantom{spac}$\mdist$ $\downarrow$\phantom{spac}}\unboldmath{} & \boldmath{}\textbf{\phantom{spa}$\mrank$ $\downarrow$\phantom{spa}}\unboldmath{} & \boldmath{}\textbf{\phantom{sp}$\mpckauc$ $\uparrow$\phantom{sp}}\unboldmath{} & \boldmath{}\textbf{\phantom{spac}CD $\downarrow$\phantom{spac}}\unboldmath{} \\
    \midrule
    \multirow{5}[2]{*}{cat\_walk} & nrICP & 77.96$\pm$94.55 & 5.72$\pm$9.65 & 70.39$\pm$15.02 & - \\
          & AN    & 14.27$\pm$18.96 & 0.41$\pm$1.04 & 97.07$\pm$1.18 & \boldmath{}\textbf{0.38$\pm$0.00}\unboldmath{} \\
          & DSR   & 48.05$\pm$80.24 & 3.05$\pm$7.62 & 82.53$\pm$11.60 & 0.41$\pm$0.01 \\
          & CC    & 53.37$\pm$93.99 & 3.90$\pm$8.83 & 80.54$\pm$14.99 & - \\
          & \textbf{OUR} & \boldmath{}\textbf{8.45$\pm$12.93}\unboldmath{} & \boldmath{}\textbf{0.20$\pm$0.56}\unboldmath{} & \boldmath{}\textbf{98.61$\pm$0.61}\unboldmath{} & \boldmath{}\textbf{0.38$\pm$0.00}\unboldmath{} \\
    \midrule
    \multirow{5}[2]{*}{horse\_gallop} & nrICP & 69.94$\pm$76.34 & 4.91$\pm$7.11 & 72.62$\pm$13.23 & - \\
          & AN    & 17.52$\pm$27.33 & 0.66$\pm$3.01 & 96.60$\pm$1.24 & \boldmath{}\textbf{0.09$\pm$0.00}\unboldmath{} \\
\bottomrule
    \end{tabular}%
  	\endgroup
	}
	\label{tab:results_individual_anim}%
\end{table}%

\begin{table*}[htbp]
  \centering
  \caption{\textbf{Comparison of \ours{} to SotA methods on correspondence accuracy and reconstruction quality on the \amaa{} dataset.}}
  \vspace{0.1cm}
  \resizebox{0.99\textwidth}{!}{
  	\begingroup
  	\setlength{\tabcolsep}{1pt}
  	\renewcommand{\arraystretch}{1.1}
    \begin{tabular}{rrrrrr|clcccc}
    \toprule
    \multicolumn{1}{c}{\textbf{\phantom{spac}sequence\phantom{spac}}} & \multicolumn{1}{l}{\textbf{model\phantom{spa}}} & \multicolumn{1}{l}{\boldmath{}\textbf{\phantom{spac}$\mdist$ $\downarrow$\phantom{spac}}\unboldmath{}} & \multicolumn{1}{l}{\boldmath{}\textbf{\phantom{spa}$\mrank$ $\downarrow$\phantom{spa}}\unboldmath{}} & \multicolumn{1}{l}{\boldmath{}\textbf{\phantom{sp}$\mpckauc$ $\uparrow$\phantom{sp}}\unboldmath{}} & \multicolumn{1}{l|}{\boldmath{}\textbf{\phantom{spac}CD $\downarrow$\phantom{spac}}\unboldmath{}} & \textbf{\phantom{spac}sequence\phantom{spac}} & \textbf{model\phantom{spa}} & \multicolumn{1}{l}{\boldmath{}\textbf{\phantom{spac}$\mdist$ $\downarrow$\phantom{spac}}\unboldmath{}} & \multicolumn{1}{l}{\boldmath{}\textbf{\phantom{spa}$\mrank$ $\downarrow$\phantom{spa}}\unboldmath{}} & \multicolumn{1}{l}{\boldmath{}\textbf{\phantom{sp}$\mpckauc$ $\uparrow$\phantom{sp}}\unboldmath{}} & \multicolumn{1}{l}{\boldmath{}\textbf{\phantom{spac}CD $\downarrow$\phantom{spac}}\unboldmath{}} \\
    \midrule
    \multicolumn{1}{c}{\multirow{5}[2]{*}{bouncing}} & \multicolumn{1}{l}{nrICP} & \multicolumn{1}{c}{130.93$\pm$111.27} & \multicolumn{1}{c}{4.11$\pm$7.76} & \multicolumn{1}{c}{43.64$\pm$17.60} & \multicolumn{1}{c|}{-} & \multirow{5}[2]{*}{march\_2} & nrICP & 187.52$\pm$182.98 & 9.33$\pm$14.65 & 39.92$\pm$23.08 & - \\
          & \multicolumn{1}{l}{AN} & \multicolumn{1}{c}{59.86$\pm$50.14} & \multicolumn{1}{c}{0.92$\pm$3.19} & \multicolumn{1}{c}{77.84$\pm$7.56} & \multicolumn{1}{c|}{0.37$\pm$0.02} &       & AN    & 120.96$\pm$166.77 & 5.03$\pm$11.33 & 64.99$\pm$18.34 & 0.30$\pm$0.02 \\
          & \multicolumn{1}{l}{DSR} & \multicolumn{1}{c}{51.09$\pm$34.89} & \multicolumn{1}{c}{0.56$\pm$0.96} & \multicolumn{1}{c}{81.57$\pm$7.00} & \multicolumn{1}{c|}{0.35$\pm$0.01} &       & DSR   & 143.05$\pm$144.18 & 5.76$\pm$10.75 & 47.77$\pm$20.80 & 0.96$\pm$0.15 \\
\end{tabular}%
  	\endgroup
	}
	\label{tab:results_individual_amaa}%
\end{table*}%

\begin{table*}[htbp]
  \centering
  \caption{\textbf{Comparison of \ours{} to SotA methods on correspondence accuracy and reconstruction quality on the \dfaust{} dataset.}}
  \resizebox{0.99\textwidth}{!}{
  	\begingroup
  	\setlength{\tabcolsep}{1pt}
  	\renewcommand{\arraystretch}{1.05}
    \begin{tabular}{rrrrrr|clcccc}
    \toprule
    \multicolumn{1}{c}{\textbf{\phantom{spac}sequence\phantom{spac}}} & \multicolumn{1}{l}{\textbf{model\phantom{spa}}} & \multicolumn{1}{l}{\boldmath{}\textbf{\phantom{spac}$\mdist$ $\downarrow$\phantom{spac}}\unboldmath{}} & \multicolumn{1}{l}{\boldmath{}\textbf{\phantom{spa}$\mrank$ $\downarrow$\phantom{spa}}\unboldmath{}} & \multicolumn{1}{l}{\boldmath{}\textbf{\phantom{sp}$\mpckauc$ $\uparrow$\phantom{sp}}\unboldmath{}} & \multicolumn{1}{l|}{\boldmath{}\textbf{\phantom{spac}CD $\downarrow$\phantom{spac}}\unboldmath{}} & \textbf{\phantom{spac}sequence\phantom{spac}} & \textbf{model\phantom{spa}} & \multicolumn{1}{l}{\boldmath{}\textbf{\phantom{spac}$\mdist$ $\downarrow$\phantom{spac}}\unboldmath{}} & \multicolumn{1}{l}{\boldmath{}\textbf{\phantom{spa}$\mrank$ $\downarrow$\phantom{spa}}\unboldmath{}} & \multicolumn{1}{l}{\boldmath{}\textbf{\phantom{sp}$\mpckauc$ $\uparrow$\phantom{sp}}\unboldmath{}} & \multicolumn{1}{l}{\boldmath{}\textbf{\phantom{spac}CD $\downarrow$\phantom{spac}}\unboldmath{}} \\
    \midrule
    \multicolumn{1}{c}{\multirow{5}[2]{*}{chicken\_wings}} & \multicolumn{1}{l}{nrICP} & \multicolumn{1}{c}{72.20$\pm$135.09} & \multicolumn{1}{c}{5.15$\pm$15.70} & \multicolumn{1}{c}{80.89$\pm$13.79} & \multicolumn{1}{c|}{-} & \multirow{5}[2]{*}{one\_leg\_jump} & nrICP & 81.13$\pm$98.03 & 2.86$\pm$5.19 & 71.68$\pm$14.61 & - \\
          & \multicolumn{1}{l}{AN} & \multicolumn{1}{c}{49.75$\pm$92.05} & \multicolumn{1}{c}{2.31$\pm$6.96} & \multicolumn{1}{c}{85.57$\pm$16.07} & \multicolumn{1}{c|}{0.37$\pm$0.08} &       & AN    & 30.65$\pm$23.71 & 0.58$\pm$0.98 & 92.47$\pm$3.48 & 0.43$\pm$0.08 \\
          & \multicolumn{1}{l}{DSR} & \multicolumn{1}{c}{282.61$\pm$130.70} & \multicolumn{1}{c}{21.85$\pm$17.43} & \multicolumn{1}{c}{6.40$\pm$2.66} & \multicolumn{1}{c|}{97.10$\pm$25.04} &       & DSR   & 47.59$\pm$99.20 & 1.30$\pm$4.72 & 88.44$\pm$8.96 & 0.36$\pm$0.07 \\
\end{tabular}%
	\endgroup
	}
	\label{tab:results_individual_dfaust}%
\end{table*}%

\begin{table}[htbp]
  \centering
  \caption{\textbf{Comparison of \ours{} to SotA methods on correspondence accuracy and reconstruction quality on \animr{} dataset.}}
  \vspace{0.1cm}
  \resizebox{0.49\textwidth}{!}{
  	\begingroup
  	\setlength{\tabcolsep}{1pt}
  	\renewcommand{\arraystretch}{1.1}
    \begin{tabular}{clcccc}
    \toprule
    \multicolumn{1}{l}{\textbf{\phantom{spac}sequence\phantom{spac}}} & \textbf{model} & \multicolumn{1}{l}{\boldmath{}\textbf{\phantom{spac}$\mdist$ $\downarrow$\phantom{spac}}\unboldmath{}} & \multicolumn{1}{l}{\boldmath{}\textbf{\phantom{spa}$\mrank$ $\downarrow$\phantom{spa}}\unboldmath{}} & \multicolumn{1}{l}{\boldmath{}\textbf{\phantom{sp}$\mpckauc$ $\uparrow$\phantom{sp}}\unboldmath{}} & \multicolumn{1}{l}{\boldmath{}\textbf{\phantom{spac}CD $\downarrow$\phantom{spac}}\unboldmath{}} \\
    \midrule
    \multirow{5}[2]{*}{cat\_walk} & nrICP & 193.93$\pm$198.01 & 21.91$\pm$26.06 & 42.42$\pm$31.89 & - \\
          & AN    & 134.49$\pm$162.04 & 13.73$\pm$20.65 & 56.01$\pm$24.78 & 0.40$\pm$0.00 \\
          & DSR   & 164.36$\pm$162.38 & 17.39$\pm$21.88 & 46.50$\pm$33.98 & 0.60$\pm$0.04 \\
          & CC    & 56.68$\pm$93.81 & 4.08$\pm$8.76 & 80.82$\pm$12.76 & - \\
          & \textbf{OUR} & \boldmath{}\textbf{11.08$\pm$15.70}\unboldmath{} & \boldmath{}\textbf{0.29$\pm$0.73}\unboldmath{} & \boldmath{}\textbf{97.97$\pm$1.02}\unboldmath{} & \boldmath{}\textbf{0.38$\pm$0.01}\unboldmath{} \\
    \midrule
    \multirow{5}[2]{*}{horse\_gallop} & nrICP & 187.99$\pm$160.25 & 21.44$\pm$25.83 & 35.52$\pm$27.97 & - \\
          & AN    & 196.12$\pm$201.63 & 21.34$\pm$26.67 & 45.15$\pm$34.19 & 0.11$\pm$0.00 \\
\bottomrule
    \end{tabular}%
	\endgroup
	}
	\label{tab:results_individual_animr}%
\end{table}%

\begin{table*}[htbp]
  \centering
  \caption{\textbf{Comparison of \ours{} to SotA methods on correspondence accuracy and reconstruction quality on \ama{} dataset.}}
  \vspace{0.1cm}
  \resizebox{0.99\textwidth}{!}{
  	\begingroup
  	\setlength{\tabcolsep}{1pt}
  	\renewcommand{\arraystretch}{1.1}
    \begin{tabular}{rrrrrr|clcccc}
    \toprule
    \multicolumn{1}{c}{\textbf{\phantom{spac}sequence\phantom{spac}}} & \multicolumn{1}{l}{\textbf{model\phantom{spa}}} & \multicolumn{1}{l}{\boldmath{}\textbf{\phantom{spac}$\mdist$ $\downarrow$\phantom{spac}}\unboldmath{}} & \multicolumn{1}{l}{\boldmath{}\textbf{\phantom{spa}$\mrank$ $\downarrow$\phantom{spa}}\unboldmath{}} & \multicolumn{1}{l}{\boldmath{}\textbf{\phantom{sp}$\mpckauc$ $\uparrow$\phantom{sp}}\unboldmath{}} & \multicolumn{1}{l|}{\boldmath{}\textbf{\phantom{spac}CD $\downarrow$\phantom{spac}}\unboldmath{}} & \textbf{\phantom{spac}sequence\phantom{spac}} & \textbf{model\phantom{spa}} & \multicolumn{1}{l}{\boldmath{}\textbf{\phantom{spac}$\mdist$ $\downarrow$\phantom{spac}}\unboldmath{}} & \multicolumn{1}{l}{\boldmath{}\textbf{\phantom{spa}$\mrank$ $\downarrow$\phantom{spa}}\unboldmath{}} & \multicolumn{1}{l}{\boldmath{}\textbf{\phantom{sp}$\mpckauc$ $\uparrow$\phantom{sp}}\unboldmath{}} & \multicolumn{1}{l}{\boldmath{}\textbf{\phantom{spac}CD $\downarrow$\phantom{spac}}\unboldmath{}} \\
    \midrule
    \multicolumn{1}{c}{\multirow{5}[2]{*}{bouncing}} & \multicolumn{1}{l}{nrICP} & \multicolumn{1}{c}{251.11$\pm$192.69} & \multicolumn{1}{c}{13.11$\pm$15.35} & \multicolumn{1}{c}{22.49$\pm$22.66} & \multicolumn{1}{c|}{-} & \multirow{5}[2]{*}{march\_2} & nrICP & 271.12$\pm$203.73 & 16.39$\pm$18.16 & 21.39$\pm$24.59 & - \\
          & \multicolumn{1}{l}{AN} & \multicolumn{1}{c}{264.71$\pm$224.26} & \multicolumn{1}{c}{15.31$\pm$17.85} & \multicolumn{1}{c}{28.91$\pm$29.76} & \multicolumn{1}{c|}{0.34$\pm$0.02} &       & AN    & 366.05$\pm$273.23 & 25.36$\pm$24.52 & 16.95$\pm$25.35 & 0.33$\pm$0.02 \\
          & \multicolumn{1}{l}{DSR} & \multicolumn{1}{c}{230.38$\pm$204.25} & \multicolumn{1}{c}{12.47$\pm$15.74} & \multicolumn{1}{c}{33.32$\pm$29.72} & \multicolumn{1}{c|}{0.54$\pm$0.04} &       & DSR   & 281.72$\pm$192.68 & 16.95$\pm$17.20 & 18.42$\pm$23.35 & 1.68$\pm$0.36 \\
\end{tabular}%
	\endgroup
	}
	\label{tab:results_individual_ama}%
\end{table*}%

\begin{table*}[htbp]
  \centering
  \caption{\textbf{Comparison of \ours{} to SotA methods on correspondence accuracy and reconstruction quality on \amaan{} dataset.}}
  \vspace{0.1cm}
  \resizebox{0.99\textwidth}{!}{
  	\begingroup
  	\setlength{\tabcolsep}{1pt}
  	\renewcommand{\arraystretch}{1.1}
    \begin{tabular}{rrrrrr|clcccc}
    \toprule
    \multicolumn{1}{c}{\textbf{\phantom{spac}sequence\phantom{spac}}} & \multicolumn{1}{l}{\textbf{model\phantom{spa}}} & \multicolumn{1}{l}{\boldmath{}\textbf{\phantom{spac}$\mdist$ $\downarrow$\phantom{spac}}\unboldmath{}} & \multicolumn{1}{l}{\boldmath{}\textbf{\phantom{spa}$\mrank$ $\downarrow$\phantom{spa}}\unboldmath{}} & \multicolumn{1}{l}{\boldmath{}\textbf{\phantom{sp}$\mpckauc$ $\uparrow$\phantom{sp}}\unboldmath{}} & \multicolumn{1}{l|}{\boldmath{}\textbf{\phantom{spac}CD $\downarrow$\phantom{spac}}\unboldmath{}} & \textbf{\phantom{spac}sequence\phantom{spac}} & \textbf{model\phantom{spa}} & \multicolumn{1}{l}{\boldmath{}\textbf{\phantom{spac}$\mdist$ $\downarrow$\phantom{spac}}\unboldmath{}} & \multicolumn{1}{l}{\boldmath{}\textbf{\phantom{spa}$\mrank$ $\downarrow$\phantom{spa}}\unboldmath{}} & \multicolumn{1}{l}{\boldmath{}\textbf{\phantom{sp}$\mpckauc$ $\uparrow$\phantom{sp}}\unboldmath{}} & \multicolumn{1}{l}{\boldmath{}\textbf{\phantom{spac}CD $\downarrow$\phantom{spac}}\unboldmath{}} \\
    \midrule
    \multicolumn{1}{c}{\multirow{5}[2]{*}{bouncing}} & \multicolumn{1}{l}{nrICP} & \multicolumn{1}{c}{136.96$\pm$109.51} & \multicolumn{1}{c}{4.32$\pm$7.37} & \multicolumn{1}{c}{40.30$\pm$16.05} & \multicolumn{1}{c|}{-} & \multirow{5}[2]{*}{march\_2} & nrICP & 197.13$\pm$181.45 & 9.97$\pm$14.91 & 35.75$\pm$21.21 & - \\
          & \multicolumn{1}{l}{AN} & \multicolumn{1}{c}{93.76$\pm$81.79} & \multicolumn{1}{c}{2.45$\pm$5.54} & \multicolumn{1}{c}{60.21$\pm$11.63} & \multicolumn{1}{c|}{0.58$\pm$0.01} &       & AN    & 131.68$\pm$159.64 & 5.50$\pm$11.81 & 55.96$\pm$17.93 & 0.58$\pm$0.01 \\
          & \multicolumn{1}{l}{DSR} & \multicolumn{1}{c}{126.07$\pm$135.63} & \multicolumn{1}{c}{4.88$\pm$10.24} & \multicolumn{1}{c}{54.68$\pm$18.18} & \multicolumn{1}{c|}{\boldmath{}\textbf{0.54$\pm$0.03}\unboldmath{}} &       & DSR   & 179.32$\pm$196.95 & 10.31$\pm$17.24 & 48.13$\pm$23.41 & \boldmath{}\textbf{0.47$\pm$0.02}\unboldmath{} \\
\end{tabular}%
  	\endgroup
	}
	\label{tab:results_individual_amaan}%
\end{table*}%

\begin{table}[htbp]
  \centering
  \caption{\textbf{Comparison of \ours{} to SotA methods on correspondence accuracy and reconstruction quality on the \cape{} dataset.}}
  \vspace{0.1cm}
  \resizebox{0.49\textwidth}{!}{
  	\begingroup
  	\setlength{\tabcolsep}{1pt}
  	\renewcommand{\arraystretch}{1.1}
    \begin{tabular}{clcccc}
    \toprule
    \textbf{\phantom{spac}sequence\phantom{spac}} & \textbf{model\phantom{s}} & \multicolumn{1}{l}{\boldmath{}\textbf{\phantom{spac}$\mdist$ $\downarrow$\phantom{spac}}\unboldmath{}} & \multicolumn{1}{l}{\boldmath{}\textbf{\phantom{spa}$\mrank$ $\downarrow$\phantom{spa}}\unboldmath{}} & \multicolumn{1}{l}{\boldmath{}\textbf{\phantom{sp}$\mpckauc$ $\uparrow$\phantom{sp}}\unboldmath{}} & \multicolumn{1}{l}{\boldmath{}\textbf{\phantom{spac}CD $\downarrow$\phantom{spac}}\unboldmath{}} \\
    \midrule
    \multirow{5}[2]{*}{shortlong\_hips} & nrICP & 82.98$\pm$74.07 & 3.83$\pm$6.00 & 65.36$\pm$19.46 & - \\
          & AN    & 34.57$\pm$40.18 & 0.85$\pm$2.72 & 90.56$\pm$5.36 & \boldmath{}\textbf{0.47$\pm$0.16}\unboldmath{} \\
          & DSR   & 29.79$\pm$50.19 & 0.84$\pm$4.18 & 93.31$\pm$6.80 & 1.14$\pm$9.13 \\
          & CC    & 32.45$\pm$54.88 & 0.93$\pm$4.27 & 91.94$\pm$12.92 & - \\
          & \textbf{OUR} & \boldmath{}\textbf{24.36$\pm$35.34}\unboldmath{} & \boldmath{}\textbf{0.56$\pm$2.75}\unboldmath{} & \boldmath{}\textbf{94.83$\pm$5.51}\unboldmath{} & 0.74$\pm$7.63 \\
    \midrule
    \multirow{5}[2]{*}{shortlong\_pose\_model} & nrICP & 107.17$\pm$106.82 & 6.29$\pm$9.81 & 57.81$\pm$23.88 & - \\
          & AN    & 140.08$\pm$186.05 & 9.94$\pm$18.20 & 55.71$\pm$23.63 & 0.32$\pm$0.03 \\
\bottomrule
    \end{tabular}%
  	\endgroup
	}
	\label{tab:results_individual_cape}%
\end{table}%

\begin{table}[htbp]
  \centering
  \caption{\textbf{Comparison of \ours{} to SotA methods on correspondence accuracy and reconstruction quality on the \inria{} dataset.}}
  \vspace{0.1cm}
  \resizebox{0.49\textwidth}{!}{
  	\begingroup
  	\setlength{\tabcolsep}{1pt}
  	\renewcommand{\arraystretch}{1.1}
    \begin{tabular}{clcccc}
    \toprule
    \textbf{\phantom{spac}sequence\phantom{spac}} & \textbf{model\phantom{spa}} & \multicolumn{1}{l}{\boldmath{}\textbf{\phantom{spac}$\mdist$ $\downarrow$\phantom{spac}}\unboldmath{}} & \multicolumn{1}{l}{\boldmath{}\textbf{\phantom{spa}$\mrank$ $\downarrow$\phantom{spa}}\unboldmath{}} & \multicolumn{1}{l}{\boldmath{}\textbf{\phantom{sp}$\mpckauc$ $\uparrow$\phantom{sp}}\unboldmath{}} & \multicolumn{1}{l}{\boldmath{}\textbf{\phantom{spac}CD $\downarrow$\phantom{spac}}\unboldmath{}} \\
    \midrule
    \multirow{5}[2]{*}{tight\_walk} & nrICP & 87.59$\pm$117.67 & 9.33$\pm$14.41 & 63.21$\pm$29.94 & - \\
          & AN    & 90.63$\pm$121.24 & 9.93$\pm$15.55 & 62.47$\pm$29.78 & 10.39$\pm$0.95 \\
          & DSR   & 125.82$\pm$150.45 & 15.33$\pm$21.39 & 55.67$\pm$30.61 & \boldmath{}\textbf{10.20$\pm$0.90}\unboldmath{} \\
          & CC    & 88.47$\pm$118.85 & 9.59$\pm$14.63 & 63.49$\pm$31.60 & - \\
          & \textbf{OUR} & \boldmath{}\textbf{77.15$\pm$120.63}\unboldmath{} & \boldmath{}\textbf{8.58$\pm$14.71}\unboldmath{} & \boldmath{}\textbf{69.58$\pm$36.47}\unboldmath{} & 10.68$\pm$1.11 \\
    \midrule
    \multirow{5}[2]{*}{wide\_knee} & nrICP & 86.18$\pm$140.62 & 10.54$\pm$19.72 & 69.05$\pm$24.68 & - \\
          & AN    & 77.68$\pm$126.27 & 8.80$\pm$16.03 & 69.26$\pm$29.90 & 12.55$\pm$1.29 \\
\bottomrule
    \end{tabular}%
  	\endgroup
	}
	\label{tab:results_individual_inria}%
\end{table}%

\begin{table}[htbp]
  \centering
  \caption{\textbf{Comparison of \ours{} to SotA methods on correspondence accuracy and reconstruction quality on a collapsing rubber horse used to stress test our method and on an additional similar sequence depicting a collapsing camel}. Our method yields much more accurate correspondences.}
  \vspace{0.1cm}
  \resizebox{0.49\textwidth}{!}{
  	\begingroup
  	\setlength{\tabcolsep}{1pt}
  	\renewcommand{\arraystretch}{1.1}
    \begin{tabular}{clcccc}
    \toprule
    \textbf{\phantom{spac}sequence\phantom{spac}} & \textbf{model} & \boldmath{}\textbf{\phantom{spac}$\mdist$ $\downarrow$\phantom{spac}}\unboldmath{} & \boldmath{}\textbf{\phantom{spa}$\mrank$ $\downarrow$\phantom{spa}}\unboldmath{} & \boldmath{}\textbf{\phantom{sp}$\mpckauc$ $\uparrow$\phantom{sp}}\unboldmath{} & \boldmath{}\textbf{\phantom{spac}CD $\downarrow$\phantom{spac}}\unboldmath{} \\
    \midrule
    \multirow{5}[2]{*}{horse\_collapse} & nrICP & 54.32$\pm$46.49 & 3.88$\pm$5.47 & 78.36$\pm$13.66 & - \\
          & AN    & 62.38$\pm$79.58 & 4.91$\pm$9.05 & 74.81$\pm$18.23 & \boldmath{}\textbf{0.13$\pm$0.01}\unboldmath{} \\
          & DSR   & 49.00$\pm$60.40 & 3.25$\pm$6.18 & 81.73$\pm$14.42 & 0.18$\pm$0.02 \\
          & CC    & 56.51$\pm$78.80 & 3.89$\pm$7.49 & 77.97$\pm$19.67 & - \\
          & \textbf{OUR} & \boldmath{}\textbf{24.69$\pm$37.43}\unboldmath{} & \boldmath{}\textbf{1.09$\pm$3.18}\unboldmath{} & \boldmath{}\textbf{93.25$\pm$6.75}\unboldmath{} & 0.14$\pm$0.01 \\
    \midrule
    \multirow{5}[2]{*}{camel\_collapse} & nrICP & 40.68$\pm$36.05 & 2.76$\pm$3.69 & 86.60$\pm$9.77 & - \\
          & AN    & 43.78$\pm$61.53 & 3.05$\pm$6.11 & 85.05$\pm$12.47 & \boldmath{}\textbf{0.16$\pm$0.01}\unboldmath{} \\
\bottomrule
    \end{tabular}%
  	\endgroup
	}
	\label{tab:results_individual_stress_test}%
\end{table}%

%% file: top_arxiv.bbl
\begin{thebibliography}{10}\itemsep=-1pt

\bibitem{Aigerman14}
N.~Aigerman, R.~Poranne, and Y.~Lipman.
\newblock {Lifted bijections for low distortion surface mappings}.
\newblock {\em ACM Transactions on Graphics}, 2014.

\bibitem{Alexa00b}
M.~Alexa, D.~Cohen-Or, and D.~Levin.
\newblock {As-rigid-as-possible shape interpolation}.
\newblock In {\em Proceedings of the ACM on Computer Graphics and Interactive
  Techniques}, 2000.

\bibitem{Arcila13}
R.~Arcila, C.~Cagniart, F.~H{\'e}troy, E.~Boyer, and F.~Dupont.
\newblock {Segmentation of temporal mesh sequences into rigidly moving
  components}.
\newblock {\em Graphical Models}, 2013.

\bibitem{Aubry11}
M.~Aubry, U.~Schlickewei, and D.~Cremers.
\newblock The wave kernel signature: A quantum mechanical approach to shape
  analysis.
\newblock In {\em International Conference on Computer Vision Workshops}, 2011.

\bibitem{Aujay07}
G.~Aujay, F.~Hetroy, F.~Lazarus, and C.~Depraz.
\newblock {Harmonic Skeleton for Realistic Character Animation}.
\newblock In {\em Eurographics}, 2007.

\bibitem{Baden18}
A.~Baden, K.~Crane, and M.~Kazhdan.
\newblock {Möbius Registration}.
\newblock {\em Computer Graphics Forum}, 2018.

\bibitem{Bednarik21}
J.~Bednarik, V.~G. Kim, S.~Chaudhuri, S.~Parashar, M.~Salzmann, P.~Fua, and
  N.~Aigerman.
\newblock {Temporally-Coherent Surface Reconstruction via Metric-Consistent
  Atlases}.
\newblock In {\em International Conference on Computer Vision}, 2021.

\bibitem{Bednarik20a}
J.~Bednar\'{\i}k, S.~Parashar, E.~Gundogdu, M.~Salzmann, and P.~Fua.
\newblock {Shape Reconstruction by Learning Differentiable Surface
  Representations}.
\newblock In {\em Conference on Computer Vision and Pattern Recognition}, 2020.

\bibitem{Berger14}
M.~Berger, A.~Tagliasacchi, L.~Seversky, P.~Alliez, J.~Levine, A.~Sharf, and
  C.~Silva.
\newblock {State of the art in surface reconstruction from point clouds}.
\newblock In {\em Eurographics}, 2014.

\bibitem{Bogo17}
F.~Bogo, J.~Romero, G.~Pons-Moll, and M.~J. Black.
\newblock {Dynamic {FAUST}: Registering Human Bodies in Motion}.
\newblock In {\em Conference on Computer Vision and Pattern Recognition}, 2017.

\bibitem{Bronstein06}
A.~M. Bronstein, M.~M. Bronstein, and R.~Kimmel.
\newblock {Generalized multidimensional scaling: A framework for
  isometry-invariant partial surface matching}.
\newblock {\em Proceedings of the National Academy of Sciences USA}, 2006.

\bibitem{Cosmo20}
L.~Cosmo, A.~Norelli, O.~Halimi, R.~Kimmel, and E.~Rodol{\`a}.
\newblock {LIMP: Learning Latent Shape Representations with Metric Preservation
  Priors}.
\newblock {\em European Conference on Computer Vision}, 2020.

\bibitem{Cosmo19}
L.~Cosmo, M.~Panine, A.~Rampini, M.~Ovsjanikov, M.~M. Bronstein, and E.~Rodola.
\newblock {Isospectralization, or how to hear shape, style, and
  correspondence}.
\newblock In {\em Conference on Computer Vision and Pattern Recognition}, 2019.

\bibitem{Cuzzolin08}
F.~Cuzzolin, D.~Mateusy, D.~Knossow, E.~Boyer, and R.~Horaud.
\newblock {Coherent Laplacian 3-D protrusion segmentation}.
\newblock In {\em Conference on Computer Vision and Pattern Recognition}, 2008.

\bibitem{Deng20b}
Z.~Deng, J.~Bednar\'{\i}k, M.~Salzmann, and P.~Fua.
\newblock {Better Patch Stitching for Parametric Surface Reconstruction}.
\newblock In {\em International Conference on 3D Vision}, 2020.

\bibitem{Deprelle19}
T.~Deprelle, T.~Groueix, M.~Fisher, V.~G. Kim, B.~C. Russell, and M.~Aubry.
\newblock {Learning Elementary Structures for 3D Shape Generation and
  Matching}.
\newblock In {\em Advances in Neural Information Processing Systems}, 2019.

\bibitem{Donati20}
N.~Donati, A.~Sharma, and M.~Ovsjanikov.
\newblock {Deep Geometric Functional Maps: Robust Feature Learning for Shape
  Correspondence}.
\newblock In {\em Conference on Computer Vision and Pattern Recognition}, 2020.

\bibitem{Fan17a}
H.~Fan, H.~Su, and L.~Guibas.
\newblock {A Point Set Generation Network for 3D Object Reconstruction from a
  Single Image}.
\newblock In {\em Conference on Computer Vision and Pattern Recognition}, 2017.

\bibitem{Fua96f}
P.~Fua.
\newblock {Model-Based Optimization: Accurate and Consistent Site Modeling}.
\newblock In {\em International Society for Photogrammetry and Remote Sensing},
  July 1996.

\bibitem{Gao17b}
L.~Gao, S.-Y. Chen, Y.-K. Lai, and S.~Xia.
\newblock {Data-Driven Shape Interpolation and Morphing Editing}.
\newblock In {\em Computer Graphics Forum}, 2017.

\bibitem{Groueix18a}
T.~Groueix, M.~Fisher, V.~Kim, B.~Russell, and M.~Aubry.
\newblock {Atlasnet: A Papier-M\^ach{\'e} Approach to Learning 3D Surface
  Generation}.
\newblock In {\em Conference on Computer Vision and Pattern Recognition}, 2018.

\bibitem{Groueix18b}
T.~Groueix, M.~Fisher, V.~G. Kim, B.~C. Russell, and M.~Aubry.
\newblock {3D-CODED : 3D Correspondences by Deep Deformation}.
\newblock In {\em European Conference on Computer Vision}, 2018.

\bibitem{Groueix19}
T.~Groueix, M.~Fisher, V.~G. Kim, B.~C. Russell, and M.~Aubry.
\newblock {Unsupervised Cycle-Consistent Deformation for Shape Matching}.
\newblock {\em Computer Graphics Forum}, 2019.

\bibitem{Halimi19}
O.~Halimi, O.~Litany, E.~Rodola, A.~M. Bronstein, and R.~Kimmel.
\newblock {Unsupervised Learning of Dense Shape Correspondence}.
\newblock In {\em Conference on Computer Vision and Pattern Recognition}, 2019.

\bibitem{Huang17f}
H.~Huang, E.~Kalogerakis, S.~Chaudhuri, D.~Ceylan, V.~G. Kim, and E.~Yumer.
\newblock {Learning Local Shape Descriptors from Part Correspondences with
  Multiview Convolutional Networks}.
\newblock {\em ACM Transactions on Graphics}, 2017.

\bibitem{Insafutdinov18}
E.~Insafutdinov and A.~Dosovitskiy.
\newblock Unsupervised learning of shape and pose with differentiable point
  clouds.
\newblock In {\em Advances in Neural Information Processing Systems}, 2018.

\bibitem{Joo15}
H.~Joo, H.~Liu, L.~Tan, L.~Gui, B.~Nabbe, I.~Matthews, T.~Kanade, S.~Nobuhara,
  and Y.~Sheikh.
\newblock {Panoptic Studio: A Massively Multiview System for Social Motion
  Capture}.
\newblock In {\em International Conference on Computer Vision}, 2015.

\bibitem{Kanazawa18b}
A.~Kanazawa, S.~Tulsiani, A.~Efros, and J.~Malik.
\newblock {Learning Category-Specific Mesh Reconstruction from Image
  Collections}.
\newblock In {\em Conference on Computer Vision and Pattern Recognition}, 2018.

\bibitem{Kazhdan13}
M.~Kazhdan and H.~Hoppe.
\newblock {Screened poisson surface reconstruction}.
\newblock {\em ACM Transactions on Graphics}, 2013.

\bibitem{Koehl14}
P.~Koehl and J.~Hass.
\newblock {Automatic Alignment of Genus-Zero Surfaces}.
\newblock {\em IEEE Transactions on Pattern Analysis and Machine Intelligence},
  36(3):466–--478, 2014.

\bibitem{Kraevoy04}
V.~Kraevoy and A.~Sheffer.
\newblock {Cross-parameterization and compatible remeshing of 3D models}.
\newblock {\em ACM Transactions on Graphics}, 2004.

\bibitem{Le16}
H.~Le, T.-J. Chin, and D.~Suter.
\newblock {Conformal Surface Alignment With Optimal Mobius Search}.
\newblock In {\em Conference on Computer Vision and Pattern Recognition}, 2016.

\bibitem{Aaron99}
A.~W.~F. Lee, D.~Dobkin, W.~Sweldens, and P.~Schr\"{o}der.
\newblock {Multiresolution Mesh Morphing}.
\newblock In {\em Proceedings of the ACM on Computer Graphics and Interactive
  Techniques}, 1999.

\bibitem{Li13e}
C.~Li and A.~B. Hamza.
\newblock {A multiresolution descriptor for deformable 3D shape retrieval}.
\newblock {\em The Visual Computer}, 29(6-8):513--524, 2013.

\bibitem{Lipman07}
Y.~Lipman, D.~Cohen-Or, and D.~Levin.
\newblock {Data-dependent MLS for faithful surface approximation}.
\newblock In {\em Eurographics}, 2007.

\bibitem{Lipman10}
Y.~Lipman and I.~Daubechies.
\newblock {Surface Comparison with Mass Transportation}, 2010.

\bibitem{Lipman09}
Y.~Lipman and T.~Funkhouser.
\newblock {M{\"o}bius Voting for Surface Correspondence}.
\newblock {\em ACM Transactions on Graphics}, 2009.

\bibitem{Loper15}
M.~Loper, N.~Mahmood, J.~Romero, G.~Pons-Moll, and M.~Black.
\newblock {SMPL: A Skinned Multi-Person Linear Model}.
\newblock {\em ACM SIGGRAPH Asia}, 34(6), 2015.

\bibitem{Ma20}
Q.~Ma, J.~Yang, A.~Ranjan, S.~Pujades, G.~Pons-Moll, S.~Tang, and M.~J. Black.
\newblock {Learning to Dress 3D People in Generative Clothing}.
\newblock In {\em Conference on Computer Vision and Pattern Recognition}, 2020.

\bibitem{Mescheder19}
L.~Mescheder, M.~Oechsle, M.~Niemeyer, S.~Nowozin, and A.~Geiger.
\newblock {Occupancy Networks: Learning 3D Reconstruction in Function Space}.
\newblock In {\em Conference on Computer Vision and Pattern Recognition}, pages
  4460--4470, 2019.

\bibitem{Memoli05}
F.~Mémoli and G.~Sapiro.
\newblock {A Theoretical and Computational Framework for Isometry Invariant
  Recognition of Point Cloud Data}.
\newblock {\em Foundations of Computational Mathematics}, 2005.

\bibitem{Niemeyer19b}
M.~Niemeyer, L.~Mescheder, M.~Oechsle, and A.~Geiger.
\newblock {Occupancy Flow: 4D Reconstruction by Learning Particle Dynamics}.
\newblock In {\em International Conference on Computer Vision}, 2019.

\bibitem{Ovsjanikov12}
M.~Ovsjanikov, M.~Ben-Chen, J.~Solomon, A.~Butscher, and L.~Guibas.
\newblock {Functional Maps: A Flexible Representation of Maps Between Shapes}.
\newblock {\em ACM Transactions on Graphics}, 31(4), 2012.

\bibitem{Pan19}
J.~Pan and K.~Jia.
\newblock {Deep Mesh Reconstruction from Single RGB Images via Topology
  Modification Networks}.
\newblock In {\em International Conference on Computer Vision}, 2019.

\bibitem{Park19c}
J.~J. Park, P.~Florence, J.~Straub, R.~A. Newcombe, and S.~Lovegrove.
\newblock {DeepSdf: Learning Continuous Signed Distance Functions for Shape
  Representation}.
\newblock In {\em Conference on Computer Vision and Pattern Recognition}, 2019.

\bibitem{Qi17a}
C.~R. Qi, H.~Su, K.~Mo, and L.~J. Guibas.
\newblock {Pointnet: Deep Learning on Point Sets for 3D Classification and
  Segmentation}.
\newblock In {\em Conference on Computer Vision and Pattern Recognition}, 2017.

\bibitem{Qi17b}
C.~R. Qi, L.~Yi, H.~Su, and L.~J. Guibas.
\newblock {Pointnet++: Deep Hierarchical Feature Learning on Point Sets in a
  Metric Space}.
\newblock In {\em Advances in Neural Information Processing Systems}, 2017.

\bibitem{Rakotosaona20}
M.-J. Rakotosaona and M.~Ovsjanikov.
\newblock {Intrinsic Point Cloud Interpolation via Dual Latent Space
  Navigation}.
\newblock In {\em European Conference on Computer Vision}, 2020.

\bibitem{Rodola14}
E.~Rodola, S.~R. Bulo, T.~Windheuser, M.~Vestner, and D.~Cremers.
\newblock {Dense non-rigid shape correspondence using random forests}.
\newblock {\em Conference on Computer Vision and Pattern Recognition}, 2014.

\bibitem{Roufosse19}
J.-M. Roufosse, A.~Sharma, and M.~Ovsjanikov.
\newblock {Unsupervised Deep Learning for Structured Shape Matching}.
\newblock In {\em International Conference on Computer Vision}, 2019.

\bibitem{Sumner04}
R.~Sumner and J.~Popovic.
\newblock {Deformation Transfer for Triangle Meshes}.
\newblock {\em ACM Transactions on Graphics}, pages 399--405, 2004.

\bibitem{Sun09c}
J.~Sun, M.~Ovsjanikov, and L.~Guibas.
\newblock {A Concise and Provably Informative Multi-Scale Signature Based on
  Heat Diffusion}.
\newblock {\em Computer Graphics Forum}, 2009.

\bibitem{Tang21}
J.~Tang, D.~Xu, K.~Jia, and L.~Zhang.
\newblock {Learning Parallel Dense Correspondence From Spatio-Temporal
  Descriptors for Efficient and Robust 4D Reconstruction}.
\newblock In {\em Conference on Computer Vision and Pattern Recognition}, 2021.

\bibitem{Varanasi10}
K.~Varanasi and E.~Boyer.
\newblock {Temporally Coherent Segmentation of 3D Reconstructions}.
\newblock In {\em 3D Data Processing, Visualization and Transmission}, 2010.

\bibitem{Vlasic08}
D.~Vlasic, I.~Baran, W.~Matusik, and J.~Popovi\'{c}.
\newblock Articulated mesh animation from multi-view silhouettes.
\newblock {\em ACM Transactions on Graphics}, 2008.

\bibitem{Wang05e}
Y.~Wang, L.~M. Lui, T.~F. Chan, and P.~M. Thompson.
\newblock {Optimization of Brain Conformal Mapping with Landmarks}.
\newblock In {\em Conference on Medical Image Computing and Computer Assisted
  Intervention}, 2005.

\bibitem{Weber14}
O.~Weber and D.~Zorin.
\newblock {Locally injective parametrization with arbitrary fixed boundaries}.
\newblock {\em ACM Transactions on Graphics}, 2014.

\bibitem{Williams19a}
F.~Williams, T.~Schneider, C.~T. Silva, D.~Zorin, J.~Bruna, and D.~Panozzo.
\newblock {Deep Geometric Prior for Surface Reconstruction}.
\newblock In {\em Conference on Computer Vision and Pattern Recognition}, 2019.

\bibitem{Winkler10}
T.~Winkler, J.~Drieseberg, M.~Alexa, and K.~Hormann.
\newblock {Multi-scale geometry interpolation}.
\newblock In {\em Computer Graphics Forum}, 2010.

\bibitem{Yang16c}
J.~Yang, J.-S. Franco, F.~H{\'e}troy-Wheeler, and S.~Wuhrer.
\newblock {Estimation of Human Body Shape in Motion with Wide Clothing}.
\newblock In {\em European Conference on Computer Vision}, 2016.

\bibitem{Yang18a}
Y.~Yang, C.~Feng, Y.~Shen, and D.~Tian.
\newblock {Foldingnet: Point Cloud Auto-Encoder via Deep Grid Deformation}.
\newblock In {\em Conference on Computer Vision and Pattern Recognition}, June
  2018.

\bibitem{You20}
Y.~You, Y.~Lou, C.~Li, Z.~Cheng, L.~Li, L.~Ma, C.~Lu, and W.~Wang.
\newblock {KeypointNet: A Large-Scale 3D Keypoint Dataset Aggregated From
  Numerous Human Annotations}.
\newblock In {\em Conference on Computer Vision and Pattern Recognition}, 2020.

\bibitem{Zuffi15}
S.~Zuffi and M.~J. Black.
\newblock {The stitched puppet: A graphical model of 3d human shape and pose.}
\newblock In {\em Conference on Computer Vision and Pattern Recognition}, 2015.

\end{thebibliography}
